\newif\iffull
\fulltrue

\iffull
  \documentclass[11pt,letterpaper]{article}
  \usepackage[notes=true,later=true]{dtrt}
  \usepackage{fullpage}
\else
  \documentclass[letter]{llncs}
  \usepackage[notes=xxx,llncssubsub]{dtrt}
\fi

\iffull
  \usepackage[backend=bibtex8,style=alphabetic,maxnames=8,maxalphanames=6,sorting=anyt]{biblatex}
  \bibliography{references}
\else
\usepackage[
  style=alphabetic,
  firstinits,
  backend=bibtex,
  maxalphanames=4,
  minalphanames=3,
  maxbibnames=15,
]{biblatex}
\AtEveryBibitem{%
  \ifentrytype{inproceedings}{%
    \clearfield{year}%
    \clearfield{pages}%
    \clearfield{booktitle}%
    \clearfield{note}%
  }{%
  }%
}
\AtEveryBibitem{%
  \ifentrytype{misc}{%
    \clearfield{year}%
  }{%
  }%
}
\AtEveryBibitem{%
  \ifentrytype{article}{%
    \clearfield{pages}%
    \clearfield{volume}%
    \clearfield{number}%
    \clearfield{month}%
  }{%
  }%
}

\DeclareFieldFormat[misc]{title}{\mkbibquote{#1\MLInpSizdot}}
\bibliography{references}
\fi

\iffull
\usepackage{amsthm}
\usepackage{thmtools}
\else

\let\endproof\relax
\usepackage{amsthm}
\fi
\usepackage{amssymb,amsfonts,amsmath}
\usepackage{bbm}
\usepackage{times}
\usepackage{microtype}
\usepackage{tikz}
	\usetikzlibrary{matrix}
	\usetikzlibrary{shapes,arrows}
	\usetikzlibrary{positioning}
\usepackage{graphicx}
\usepackage{float}
\usepackage{verbatim}
\usepackage{array}
\usepackage{multirow}
\usepackage{latexsym}
\usepackage{paralist}
\usepackage{enumitem}
  \setlist[itemize]{leftmargin=*}
  \setlist[enumerate]{leftmargin=*}
\usepackage[nameinlink,capitalise]{cleveref}
  \crefname{step}{Step}{Steps}
  \creflabelformat{equation}{#2#1#3}
\usepackage{dsfont}
\usepackage{url}
\usepackage{braket}
\usepackage{mathrsfs}
\usepackage{color}
\usepackage{soul}
\usepackage{mdframed}
\usepackage{complexity}
\usepackage{mathtools}
\usepackage{bbm}
\usepackage{setspace}
\usepackage{tabu}
\usepackage{multirow}
\usepackage{adjustbox}
\usepackage{booktabs}
\iffull
\usepackage[margin=4mm,small,labelfont=bf]{caption}
\fi

\newtheorem{itheorem}{Theorem}

\iffull
\newtheorem{theorem}{Theorem}[section]
\newtheorem{corollary}[theorem]{Corollary}

\newtheorem{definition}[theorem]{Definition}
\newtheorem{lemma}[theorem]{Lemma}
\newtheorem{claim}[theorem]{Claim}

\fi

\theoremstyle{definition} 
\iffull

\newtheorem{remark}[theorem]{Remark}
\fi

\allowdisplaybreaks
\newcommand{\FormatAuthor}[3]{
\begin{tabular}{c}
#1 \\ {\small\texttt{#2}} \\ {\small #3}
\end{tabular}
}
\newcommand{\doclearpage}{%
\iffull
\clearpage
\fi
}
\newcommand{\DoQuote}[1]{``#1''}
\newcommand{\defemph}[1]{\textbf{\emph{#1}}}
\newcommand{\keywords}[1]{\bigskip\par\noindent{\footnotesize\textbf{Keywords\/}: #1}}

\tikzstyle{source} = [rectangle, draw, fill=gray!20, 
text width=10em, text centered, minimum height=8em]

\tikzstyle{protocol} = [rectangle, draw, fill=white!20, 
text width=10em, text centered, minimum height=8em]

\tikzstyle{transformation} = [rectangle, draw, fill=white!20, 
text width=8em, text centered, rounded corners, minimum height=6em]

\tikzstyle{line} = [draw, -latex']

\DeclareSymbolFont{bbold}{U}{bbold}{m}{n}

\newcommand{\abs}[1]{\left\lvert{#1}\right\rvert}

\newcommand{\N}{\mathbb{N}}
\newcommand{\bits}{\{0,1\}}
\newcommand{\SummationDomain}{S}
\newcommand{\Functions}[2]{\mathsf{Func}(#1,#2)}
\newcommand{\FunDomain}{X}
\newcommand{\FunRange}{Y}

\newcommand{\Security}{\lambda}

\newcommand{\vect}[1]{\mathbf{#1}}
\newcommand{\concat}{\|}

\newcommand{\Malicious}[1]{\widetilde{#1}}

\newcommand{\Relation}{R}
\newcommand{\Language}{L}
\newcommand{\Instance}{\mathbbm{x}}
\newcommand{\InstanceSize}{n}
\newcommand{\Witness}{\mathbbm{w}}
\newcommand{\GetLanguage}[1]{\Language(#1)}
\newcommand{\WitnessesOf}[3]{\mathsf{Witness}^{#3}(#1,#2)}

\newcommand{\prob}[1]{\mathrm{Pr}\left[\, #1\,\right]}
\newcommand{\negl}[1]{\mathrm{negl(#1)}}
\newcommand{\probOverExp}[2]{\mathrm{Pr}_{\scriptscriptstyle#1}\left[ #2\right]}
\newcommand{\RandomExperiment}{\mathrm{EXP}}
\newcommand{\SupportOf}[1]{\mathsf{Supp}(#1)}
\newcommand{\Sampled}[1]{#1^{\star}}

\newcommand{\RanOracleSec}[1]{\mathcal{U}_{#1}}
\newcommand{\ROInstance}{u}
\newcommand{\IndexEncoding}[2]{e_{#1,#2}}
\newcommand{\QuerySet}{Q}
\newcommand{\RandomOracle}{\mathcal{U}}
\newcommand{\GeneralOracle}{\mathcal{O}}
\newcommand{\GeneralOracleInstance}{f}
\newcommand{\OracleInputSize}{\ell_{\mathsf{in}}}
\newcommand{\EmptyOracle}{O_\bot}
\newcommand{\OracleQuery}{q}
\newcommand{\OracleAnswer}{\mathsf{a}}
\newcommand{\ROInstanceWithFirstWitness}{\ROInstance_{(0)}}

\newcommand{\Insert}[2]{\InsertOperation(#1,#2)}
\newcommand{\Recover}[3]{\RecoverOperation(#1,#2,#3)}
\newcommand{\InsertOperation}{\mathsf{Insert}}
\newcommand{\RecoverOperation}{\mathsf{Recover}}
\newcommand{\WitnessDomain}{\bits^{t(\InstanceSize)}}
\newcommand{\BitsForWitness}{s}
\newcommand{\WitnessIndexRange}[1]{[t(#1)]}
\newcommand{\URel}{\Relation_{\mathsf{U}}}

\newcommand{\IARG}{\mathsf{ARG}}

\newcommand{\IARGProver}{\mathcal{P}}
\newcommand{\IARGVerifier}{\mathcal{V}}
\newcommand{\IARGCompleteness}{\alpha}
\newcommand{\IARGSoundness}{\beta}
\newcommand{\IARGSoundnessAdvQueryBound}{\mathsf{q}_{\Malicious{\IARGProver}}}
\newcommand{\IARGSoundnessAdvTimeBound}{\mathsf{t}_{\Malicious{\IARGProver}}}
\newcommand{\IARGSoundnessAdv}{\Malicious{\IARGProver}}
\newcommand{\IARGProverQuery}{\mathsf{pq}}
\newcommand{\IARGVerifierQuery}{\mathsf{vq}}
\newcommand{\IARGVerifierTime}{\mathsf{vt}}
\newcommand{\IARGVerifierRandomness}{\rho}
\newcommand{\IARGVerifierRandomnessComplexity}{\mathsf{vr}}
\newcommand{\IARGProverTime}{\mathsf{pt}}
\newcommand{\IARGCompletenessAdv}{\mathcal{A}}
\newcommand{\IARGProverRandomness}{\zeta}
\newcommand{\IARGProverRandomnessComplexity}{\mathsf{pr}}
\newcommand{\IARGVMsgForFirstWitness}[1]{a^{(0)}_{#1}}
\newcommand{\IARGPMsgForFirstWitness}[1]{b^{(0)}_{#1}}
\newcommand{\IARGDecisionForFirstWitness}{\IARGDecision^{(0)}}
\newcommand{\IARGProverComm}{\mathsf{pc}}
\newcommand{\IARGDecision}{\mathsf{d}}
\newcommand{\IARGRoundComplexity}{\mathsf{k}}
\newcommand{\IARGProverMessage}{b}
\newcommand{\IARGVerifierMessage}{a}
\newcommand{\IARGTranscript}{\mathsf{tr}}
\newcommand{\VQueryPredicate}[4]{\mathrm{Q_V}(#1,#2,#3,#4)}

\newcommand{\ZKSim}{\mathcal{S}}
\newcommand{\IARGZeroKnowledge}{z}
\newcommand{\OracleAlgorithmView}[5]{\mathsf{View}_{#1}(#2,#3,#4,#5)}
\newcommand{\ZKSimTime}{\mathsf{st}}
\newcommand{\ZKSimQuery}{\mathsf{sq}}
\newcommand{\ZKAdversaryQuery}{\mathsf{aq}}
\newcommand{\ZKAdversaryTime}{\mathsf{at}}
\newcommand{\ZKAdversary}{\Malicious{\IARGVerifier}}
\newcommand{\ZKAdversaryRandomness}{r}
\newcommand{\out}{\mathsf{out}}
\newcommand{\QueryAnswerList}{\mu}
\newcommand{\ViewFunction}[1]{\mathsf{View}_{#1}}

\newcommand{\PCPProver}{\mathcal{P}_{\mathsf{PCP}}}
\newcommand{\PCPVerifier}{\mathcal{V}_{\mathsf{PCP}}}
\newcommand{\PCPProof}{\pi} 
\newcommand{\PCPProofLength}{\ell}
\newcommand{\PCPRandomness}{r}
\newcommand{\PCPRandomnessComplexity}{\mathsf{r}}
\newcommand{\PCPQueryComplexity}{\mathsf{q}_{\mathsf{PCP}}}
\newcommand{\PCPQueryIndices}{I}
\newcommand{\PCPVerifierTime}{\mathsf{vt}_\mathsf{PCP}}
\newcommand{\PCPCompletenessError}{\gamma_{\mathsf{PCP}}}
\newcommand{\PCPKnowledgeSoundness}{\kappa_{\mathsf{PCP}}}
\newcommand{\PCPExtractor}{\mathcal{E}_{\mathsf{PCP}}}
\newcommand{\PCPExtractorTime}{t_{\mathsf{E}}}
\newcommand{\PCPSimulator}{\mathcal{S}_{\mathsf{PCP}}}

\newcommand{\HashFamily}{\mathsf H}
\newcommand{\HashGen}{\HashFamily.\mathsf{Gen}}
\newcommand{\HashEval}{H}
\newcommand{\HashKey}{\mathsf{hk}}
\newcommand{\HashCollision}{\varepsilon_{\mathsf{cr}}}

\newcommand{\VC}{\mathsf{VC}}
\newcommand{\VCGen}{\VC.\mathsf{Gen}}
\newcommand{\VCCommit}{\VC.\mathsf{Commit}}
\newcommand{\VCOpen}{\VC.\mathsf{Open}}
\newcommand{\VCCheck}{\VC.\mathsf{Check}}
\newcommand{\VCParams}{\mathsf{pp}_{\mathsf{vc}}}
\newcommand{\VCLength}{N}
\newcommand{\VCMessage}{m}
\newcommand{\VCVector}{\vect{\VCMessage}}
\newcommand{\VCCommitment}{\mathsf{cm}}
\newcommand{\VCState}{\mathsf{aux}}
\newcommand{\VCOpening}{\mathsf{pf}}
\newcommand{\VCAdv}{\mathcal{A}}
\newcommand{\VCAnswers}{\mathsf{ans}}
\newcommand{\VCBinding}{\beta_{\mathsf{vc}}}
\newcommand{\VCHiding}{\eta_{\mathsf{vc}}}

\newcommand{\IARGWitnessHiding}{\varepsilon_{\scriptscriptstyle\mathsf{WH}}}
\newcommand{\WitnessFindingError}{\varepsilon_{\scriptscriptstyle\mathsf{find}}}
\newcommand{\WHSampleAdversary}{\mathcal{A}_{\scriptscriptstyle\mathsf{WH},1}}
\newcommand{\WHOutAdversary}{\mathcal{A}_{\scriptscriptstyle\mathsf{WH},2}}
\newcommand{\WHAdversaryQuery}{\mathsf{aq}}
\newcommand{\WHAdversaryTime}{\mathsf{at}}

\newcommand{\GenericOracleAlgorithm}{{A}}
\newcommand{\GenericOracleAlgorithmInput}{z}
\newcommand{\GenericOracleAlgorithmInputSet}{Z}
\newcommand{\GenericQueryBound}{\mathsf{q}}
\newcommand{\GenericTimeBound}{\mathsf{t}}
\newcommand{\GenericOracleAlgorithmRandomness}{r}
\newcommand{\GenericOracleAlgorithmRandomnessComplexity}{\mathsf{ar}}
\newcommand{\GenericPredicate}{\phi}

\newcommand{\aux}{\mathsf{aux}}
\newcommand{\Distribution}{\mathcal{D}}
\newcommand{\DistDomain}{\mathcal{X}}

\newcommand{\DEBATE}{\mathsf{DEBATE}}
\newcommand{\DebateProver}{\IARGProver_1}
\newcommand{\DebateOpponent}{\IARGProver_2}
\newcommand{\Judge}{\mathcal{J}}

\newcommand{\SigScheme}{\mathsf{SIG}}
\newcommand{\SigGen}{\mathsf{Gen}}
\newcommand{\SigSign}{\mathsf{Sign}}
\newcommand{\SigVerify}{\mathsf{Verify}}
\newcommand{\SigSK}{\mathsf{sk}}
\newcommand{\SigVK}{\mathsf{vk}}
\newcommand{\KeyQuery}{\mathsf{key}}
\newcommand{\SignedFunction}{\mathsf{Signed}}
\newcommand{\Sig}{\sigma}
\newcommand{\SigMessage}{m}
\newcommand{\SigAdv}{\mathcal{A}}
\newcommand{\SigUnforgeability}{\varepsilon_{\mathsf{sig}}}

\newcommand{\SignedOracle}{\widehat{\GeneralOracle}}
\newcommand{\SignedOracleInstance}{\widehat{\GeneralOracleInstance}}

\newcommand{\KSAdversaryQuery}{\mathsf{aq}}
\newcommand{\KSAdversaryTime}{\mathsf{at}}
\newcommand{\KSExtractor}{\mathcal{E}}
\newcommand{\IARGKnowledgeSoundness}{\kappa}
\newcommand{\KSExtractorTime}{\mathsf{et}}
\newcommand{\KSExtractorQuery}{\mathsf{eq}}

\begin{document}

\title{
Can AI Oversight Be Zero Knowledge? 
}
\iffull
\author{
  \begin{tabular}[h!]{ccc}
  	\FormatAuthor{Alessandro Chiesa}{alessandro.chiesa@epfl.ch}{EPFL}
	& \FormatAuthor{Ziyi Guan}{ziyiguan@mit.edu}{MIT}
	& \FormatAuthor{Burcu Yıldız}{burcu.yildiz@epfl.ch}{EPFL}
  \end{tabular}
}
\date{\today}
\else
\author{}
\date{}
\institute{}
\fi

\maketitle

\begin{abstract}

AI systems increasingly produce outputs from confidential data, such as a fitness-for-duty assessment from medical records or the predicted properties of a drug candidate from its secret structure. It is important to verify that such outputs are correct without revealing the underlying data. A recent line of work studies verification of AI outputs via interactive proofs and debate for oracle-aided computation, where correctness may depend on an oracle such as human judgment, a physical experiment, or the web. These works focus on verification by a verifier that runs much faster than the computation. However, such efficient verification is impossible for general oracle-aided computation, and these works therefore rely on additional assumptions. We focus instead on privacy: allowing the verifier to run in time polynomial in the computation, we ask whether interactive arguments for oracle-aided computation can be zero knowledge, so that the verifier learns nothing about the confidential data beyond the correctness of the output.

We prove that, in general, they cannot. In the random oracle model, there are no zero-knowledge proofs for all oracle-aided computations, even if both the prover and the verifier are allowed to run much longer than the computation itself. The impossibility extends to debate, a canonical model for scalable oversight.

On the positive side, we show that if the oracle attaches a cryptographic signature to each of its answers, then every oracle-aided computation can be verified in zero knowledge with an efficient prover and verifier, assuming only collision-resistant hash functions. Beyond privacy, this also gives an alternative approach to scalable oversight that relies neither on an honest opponent, as in debate, nor on the robustness of the computation, as in prior single-prover protocols.

\keywords{zero knowledge; relativized arguments; debate; oracle-aided computation;  scalable oversight}

\vfill
\end{abstract}


\iffull
\clearpage
\setcounter{tocdepth}{2}
\begin{spacing}{0.95}
{\small\tableofcontents}
\end{spacing}
\clearpage
\else
\fi

\doclearpage
\section{Introduction}
\label{sec:introduction}

AI systems now draft medical assessments \cite{SinghalEtAl23,MoorEtAl23,ThirunavukarasuEtAl23,PadovanCPNPSCDMNLMFP24,VellaSFVSPRIZDR26}, design candidate molecules \cite{ZhavoronkovEtAl19,StokesEtAl20,WongEtAl24,XuRWEA25,JayatungaABJM24}, and produce analyses that others rely on \cite{NoyZ23,BrynjolfssonLR25}.  Although much of the data behind these outputs should remain confidential, it has repeatedly been exposed in practice. DeepSeek left a database with over a million log entries, including chat histories and API secrets, publicly accessible without authentication \cite{Nagli25}. Shared ChatGPT conversations and hundreds of thousands of shared Grok conversations became publicly searchable, exposing sensitive personal information \cite{Shanklin25,MartinI25}. Data brokers reportedly sold searchable chatbot transcripts harvested through browser extensions, including patient information entered by healthcare workers \cite{Claburn26}. Researchers have also shown how a crafted email or web page could trick a vulnerable AI assistant into leaking sensitive information \cite{AimLabs25,Lakshmanan26}.

These incidents motivate stronger privacy protections for data entrusted to AI systems. At the same time, there is growing demand for AI outputs that can be checked efficiently for correctness \cite{AZWG21,ICA18,BCIP24,KCELMB24}. Even if the leaks described above are prevented, the verification process may itself expose sensitive data. Two notable examples illustrate this.
\begin{itemize}[noitemsep]
  \item \emph{Occupational health.} Consider an AI system that drafts a fitness-for-duty assessment from an employee's confidential medical records, which an occupational physician then reviews. The employer, or a licensing body, needs to know whether the employee is \DoQuote{fit to work at heights}, but the employee should not have to hand over their medical records. Yet reaching that conclusion requires a physician's judgment about those very records.
  \item \emph{Drug discovery.} Before committing to an AI-designed drug candidate, a prospective partner wants evidence that the molecule has the claimed properties. Checking them requires a laboratory \emph{assay}: an experiment, such as a binding or toxicity test, that measures a property of the molecule. Molecular optimization treats the assay as a black-box oracle that the design algorithm queries at a cost \cite{GaoFSC22}. But running the assay independently requires a physical sample of the molecule, and a sample can be analyzed to recover its structure, which is the trade secret.
\end{itemize}

\parhead{Scalable oversight and oracle-aided computation}
In both examples, checking the AI's output requires an external judgment or measurement: the physician's judgment, or the laboratory assay. The same dependence on external resources is the defining feature of \emph{scalable oversight}, which asks how a human or a cheaper model can check an AI system's output without carrying out the full task itself \cite{CSA18,LKEMML18,IM25}. Following this line of work \cite{BCIP24,ChenKX26}, we model an external judgment or resource as a function $\GeneralOracleInstance$, called an \emph{oracle}, which the computation can query in a black-box way.

In this oracle-aided setting, scalable oversight carries out its checks through protocols that build on \emph{interactive proofs} \cite{GoldwasserMR89}, in which a \emph{prover} exchanges messages with an efficient \emph{verifier} to establish a claim. The verifier accepts true claims from an honest prover with high probability, while any cheating prover has only a small chance of convincing it to accept a false claim. Here, the AI system acts as the prover: it claims that its output $\Instance$ is accepted by a public oracle-aided computation $D$, that is, $D^{\GeneralOracleInstance}(\Instance)=1$. Both prover and verifier can query the same oracle. The goal is \emph{succinct verification}: the verifier checks this claim reliably using much less time and fewer oracle queries than the computation $D$.

\parhead{From succinctness to privacy}
We ask whether the verifier can check such a claim while the underlying data remains confidential. The prover additionally knows confidential data (the \emph{witness} $\Witness$) that the verifier does not receive, and the claim becomes \DoQuote{there exists $\Witness$ such that $D^{\GeneralOracleInstance}(\Instance,\Witness)=1$}. In our examples, $\Witness$ is the medical records or the molecular structure, and $\Instance$ is the published conclusion, together with information identifying the confidential object it concerns. The employer or business partner should be convinced that a valid witness supports the claim, without learning anything about the confidential data. This privacy requirement can be formalized using \emph{zero knowledge} \cite{GoldwasserMR89}: the verifier can check the correctness of the claim without learning anything about the hidden data beyond what the claim itself reveals.

In the occupational-health example, the AI is the prover and the employer or licensing body is the verifier. The computation $D$ takes the medical records $\Witness$ and the conclusion $\Instance$, asks the physician, the oracle, the medical questions needed to check the assessment, and accepts if the conclusion is supported. In drug discovery, the AI is the prover and the prospective partner is the verifier. The computation $D$ takes the molecule $\Witness$ and the claimed properties $\Instance$, runs the laboratory assay, the oracle, on the molecule, and accepts if the measurements match the claim.

\parhead{Our model: zero-knowledge oversight}
Ideally, checking an AI's output would be both succinct and zero knowledge. However, \cite{BarbaraCG25} establish limitations on succinct verification for oracle-aided computations, even without a privacy requirement. But an employer or a business partner would be willing to spend more time checking a conclusion if that is what protecting the underlying data requires. We therefore relax the succinctness requirement and ask:
\begin{center}
\textbf{Can every oracle-aided computation be verified in zero knowledge,\\ even if we do not require succinct verification?}
\end{center}

\subsection{Our results}
\label{sec:our-results}

We answer this question in the negative: not every oracle-aided computation can be verified in zero knowledge, even if we do not require succinct verification. To work around this impossibility, we construct a protocol for succinct zero-knowledge verification under standard cryptographic assumptions by modifying the oracle to provide digital signatures on its answers. This requires the oracle, or a party operating it, to sign its answers, an assumption that we discuss below.

\parhead{The impossibility result}
Our impossibility result holds in the \emph{random oracle model}, where the oracle is a uniformly random function. This is the standard model for relativized arguments \cite{ChiesaT10,ChenCS22}, and the model in which the limitation of \cite{BarbaraCG25} on succinct verification is proved. We prove it even after weakening both the correctness and privacy requirements:
\begin{inparaenum}[(i)]
\item we only require the verifier to reject false claims from polynomial-time cheating provers, as in \emph{relativized arguments};
\item we weaken privacy to \emph{witness hiding}: the verifier may learn partial information about the confidential data, but should not be able to recover it in full or find alternative data that would pass the check for the same claim. Since zero knowledge implies witness hiding, this rules out zero-knowledge verification as well.
\end{inparaenum}

\begin{itheorem}[informal version of \Cref{thm:wh-lower-bound-rom}]
\label{ithm:wh}
There exists an oracle-aided computation in the random oracle model that has no witness-hiding relativized argument, even when succinct verification is not required.
\end{itheorem}

Our lower bound concerns single-prover verification, the setting studied by \cite{ChenKX26} for oracle-aided scalable oversight. Recall that \cite{BarbaraCG25} rules out succinct single-prover verification for general oracle-aided computations. \cite{ChenKX26} circumvents this limitation by assuming the computation is \emph{robust}: its output tolerates a small fraction of incorrect oracle answers. This allows the verifier to check the computation using only a few oracle queries. But using a similar assumption in zero knowledge would in addition require the verifier's queries and the answers it receives to reveal nothing about the hidden witness. This need not hold in general for computations we care about. For example, a query asking whether a molecule with a specified structure binds to a target reveals that entire structure to the verifier.

Another canonical model for scalable oversight is \emph{debate}, where two provers compete to convince a verifier and correctness relies on the strong assumption that one prover acts honestly \cite{ICA18,BCIP24}. This additional assumption allows debate to circumvent the limitation of \cite{BarbaraCG25} on succinct verification. We can define privacy for debate in two ways, depending on which provers receive the secret witness. If both provers receive it, no privacy is possible, since either prover may be dishonest and send the witness to the verifier. If only the honest prover receives it, the weakest privacy requirement is that the opponent and the verifier cannot recover the witness even when both follow the protocol (a malicious opponent or verifier can in particular follow the protocol, and zero knowledge in particular hides the witness). Our impossibility rules out even this requirement, and hence every stronger one that an application may need. See \Cref{sec:techniques-debate} for details.

We therefore need a different approach to obtain zero knowledge for general oracle-aided computations.

\parhead{Construction with signed oracles}
\cite{ChiesaT10} introduced the \emph{signed random oracle model}, in which the random oracle returns each answer together with a digital signature on the query and its answer, and constructed relativized arguments with succinct verification in this model. We give a general oracle the same interface. A digital signature scheme lets the holder of a secret signing key $\SigSK$ produce a signature $\sigma$ on any string $m$. Anyone holding the matching public verification key $\SigVK$ can then check that $\sigma$ is a valid signature on $m$, and without $\SigSK$ nobody can produce a valid signature on a new string. An oracle $\GeneralOracleInstance$ answers a query $q$ with an answer $a$:
\[
  \GeneralOracleInstance(q) = a \enspace.
\]
Its signed version $\SignedOracleInstance$ returns the same answer together with a signature on the query and the answer:
\[
  \SignedOracleInstance(q) = (a, \sigma)
  \text{ where }
  \sigma \gets \SigSign\bigl(\SigSK, (q, a)\bigr)\enspace.
\]
The signature certifies that $a$ is the oracle's answer to $q$, and anyone holding $\SigVK$ can check this without querying $\GeneralOracleInstance$.

With this interface, we obtain zero-knowledge arguments with efficient proving and succinct verification for all oracle-aided computations.

\begin{itheorem}[informal version of \Cref{thm:zk-signed-oracle}]
\label{ithm:signing}
Assume the existence of collision-resistant hash functions. For every oracle $\GeneralOracleInstance$ and every oracle-aided computation, there exists a zero-knowledge single-prover argument with an efficient prover and succinct verification using the signed oracle $\SignedOracleInstance$.
\end{itheorem}
The signed oracle requires a party that holds the signing key and signs the oracle's answers. When the oracle is a digital resource, such as a database or a model, this party can be trusted hardware that evaluates the oracle and signs its answers, as studied in cryptography \cite{Katz07,PassST17}, in secure systems and databases \cite{McKeenABRSSS13,CostanD16,BajajS14,PriebeVC18}, and in machine learning \cite{MoHKMPK21,Rowstron26}. When the oracle is a person, as the physician in the occupational-health example, the person signs their judgments. When the oracle is a physical experiment, as the assay in the drug-discovery example, the laboratory that runs the experiment signs its results. The signer sees the queries, as the oracle does, and is trusted to sign only genuine answers and to keep its signing key private.

Beyond privacy, our construction provides a new alternative for scalable oversight. The limitation of \cite{BarbaraCG25} on succinct verification has so far been circumvented by assuming an honest opponent, in debate \cite{ICA18,BCIP24}, or a robust computation \cite{ChenKX26}. Signed oracles replace these assumptions by trust in a signer for the oracle's answers. \cite{ChiesaT10} showed that signatures suffice for succinct verification in the signed random oracle model; we show that they also suffice for zero knowledge, for every oracle relative to which collision-resistant hash functions exist.

\subsection{Related work}
\label{sec:related-work}

\parhead{Private verification of AI systems}
Cryptographic verification of machine-learning computations includes zero-knowledge proofs of neural-network inference, from convolutional networks \cite{LiuXZ21} to large language models \cite{SunLZ24}, proof systems optimized for machine-learning operations \cite{ChenWSK24}, and zero-knowledge proofs of training \cite{GargGJMMPW23,AbbaszadehPKP24}. Other works establish that privacy and verifiability are compatible when the model is audited for certain criteria if the audit is specified as an ordinary computation: ZkAudit verifies computations over private models and training data \cite{WaiwitlikhitSSHK24} and OATH certifies fairness of decisions while protecting model and user data \cite{FranzeseEtAl25}. On the other hand, our question concerns checks whose validity also depends on external oracle answers.

\parhead{Scalable oversight}
Scalable oversight concerns reliable supervision or evaluation of an AI model, even when the task/model exceeds a human judge's capabilities. Works such as \cite{CSA18,LKEMML18,BurnsEtAl24} focus on scalable supervision in training. On the other hand, \cite{BowmanEtAl22} propose an empirical methodology for measuring progress on scalable oversight. \cite{ICA18} propose debate as an alternative, using competing provers to help a weaker verifier. \cite{BCIP24} formalize doubly-efficient debate for oracle-aided tasks, building on doubly-efficient interactive proofs \cite{GoldwasserKR15}. \cite{BPHI25} examine the relevance of the debate model to alignment safety and identify the assumptions needed. \cite{IM25} explain the role of relativization when verification depends on human judgments or other external resources. \cite{Bar20} identifies the obfuscated-arguments problem for debate and suggests that \DoQuote{small} arguments are needed. Later work \cite{BCIP25} addresses obfuscation by prover-estimator debate. \cite{ChenKX26} show that we can consider the simpler model with a single prover if the oracle-aided computation is robust or the oracle is low-degree. These works principally address correctness and the cost of oversight, rather than simulation-based privacy for a confidential witness. We keep their oracle-aided formulation but ask whether the witness can remain hidden even when the verifier is allowed a polynomial overhead.

\parhead{Relativized arguments}
The study of relativized arguments is also motivated by proofs that verify other proofs, as in incrementally verifiable computation and proof-carrying data \cite{Valiant08,ChiesaT10,BCCT12}. The issue is that the computation being proved may invoke the same oracle as the proof system. Positive constructions exploit signed responses \cite{ChiesaT10,MallerMZ25}, low-degree encodings \cite{ChenCS22}, or arithmetized oracle access \cite{ChenCGOS23}. \cite{ChiesaGSY24} analyze the security of recursive proof composition in oracle models. Oracle-relative formulations also arise in the study of white-box security for the Fiat--Shamir transformation \cite{ArnonY25}.

On the negative side, \cite{ChiesaL20} establish limitations of probabilistically checkable proofs and interactive oracle proofs in relativized worlds. Most directly, \cite{BarbaraCG25} rule out succinct relativized arguments in the random oracle model \cite{BellareR93}, even with interaction and computational soundness. Our lower bound is complementary: it imposes no succinctness requirement and rules out even witness hiding \cite{FeigeS90}, a requirement weaker than zero knowledge \cite{GoldwasserMR89} (\Cref{lem:zk-implies-wh}).

\doclearpage
\section{Technical overview}
\label{sec:techniques}

We explain the main ideas behind our results. \Cref{sec:techniques-lower-bound} explains the proof of the lower bound (\Cref{ithm:wh}), \Cref{sec:techniques-debate} explains why the lower bound extends to debate, and \Cref{sec:techniques-signed-oracles} explains the construction with signed oracles (\Cref{ithm:signing}).

\subsection{The lower bound}
\label{sec:techniques-lower-bound}

Recall that our goal is to construct a \emph{relativized interactive argument} for an oracle-aided computation $D$ with access to an arbitrary oracle $\GeneralOracleInstance$. The prover receives $(\Instance,\Witness)$, while the verifier receives only $\Instance$; both can query $\GeneralOracleInstance$. The prover must convince the verifier that there exists a private witness $\Witness$ such that $D^{\GeneralOracleInstance}(\Instance,\Witness)=1$. Correctness requires the verifier to accept true claims from an honest prover (\emph{completeness}) and reject false claims from computationally bounded cheating provers (\emph{soundness}), with high probability. Zero knowledge requires the interaction to reveal nothing about $\Witness$. Our lower bound rules out even \emph{witness hiding}, the weaker requirement of \Cref{sec:our-results}: the verifier may learn partial information about a witness, as long as what it learns is not enough to recover an entire valid witness. Since zero knowledge implies witness hiding (\Cref{lem:zk-implies-wh}), ruling out witness hiding rules out zero knowledge as well, and we speak of witness hiding from here on.

Our impossibility result comes from a conflict between correctness and witness hiding. We choose $\GeneralOracleInstance$ to be a \emph{random oracle}: a function sampled uniformly at random and then kept fixed. A protocol for arbitrary oracles would have to work for this family too. We construct a computation $C_\star$ for which any correct protocol forces the verifier to learn a valid witness with constant probability. A witness cannot be found from the public input and oracle access alone, so a protocol that reveals one to the verifier gives it something it could never have computed itself. This is exactly what witness hiding forbids.

\begin{figure}[!b]
\centering
\begin{tikzpicture}[
  cell/.style={draw=gray!60, minimum width=7mm, minimum height=5.5mm, inner sep=0pt, anchor=center, font=\small},
  note/.style={font=\small, align=left, inner sep=1pt},
]
\matrix (T) [matrix of nodes, nodes={cell}, column sep=-\pgflinewidth, row sep=-\pgflinewidth,
  row 4/.style={nodes={fill=green!20}}] {
 1 & 0 & $\cdots$ & 1 \\
 0 & 1 & $\cdots$ & 0 \\
 $\vdots$ & $\vdots$ & & $\vdots$ \\
 0 & 0 & $\cdots$ & 0 \\
 $\vdots$ & $\vdots$ & & $\vdots$ \\
 1 & 1 & $\cdots$ & 0 \\
};
\node[above=1mm of T-1-1, font=\small] {$i=1$};
\node[above=1mm of T-1-2, font=\small] {$2$};
\node[above=1mm of T-1-3, font=\small] {$\cdots$};
\node[above=1mm of T-1-4, font=\small] {$m$};
\node[left=2mm of T-1-1, font=\small] {$0\cdots00$};
\node[left=2mm of T-2-1, font=\small] {$0\cdots01$};
\node[left=2mm of T-4-1, font=\small] {$\Witness$};
\node[left=2mm of T-6-1, font=\small] {$1\cdots11$};
\draw[red!70!black, very thick] (T-4-2.north west) rectangle (T-4-2.south east);
\node[note, right=8mm of T-4-4, anchor=west] (wnote) {all-zero row: its label $\Witness$ is a witness};
\draw[->] (wnote.west) -- (T-4-4.east);
\end{tikzpicture}
\caption{The table of bits read by the hard computation $C_\star$: $2^m$ rows, one per candidate witness, and $m$ columns. A witness is the label of an all-zero row (shaded); the boxed entry is what one query reads.}
\label{fig:table}
\end{figure}

\parhead{A hard computation}
We use the computation underlying the relation of \cite{BarbaraCG25}. Fix the public input $\Instance=0^n$ and let candidate witnesses be $m$-bit strings. The computation accepts exactly when
\begin{equation}
\label{eq:overview-hard-computation}
C_\star^{\GeneralOracleInstance}(0^n,\Witness)=1
\quad\Longleftrightarrow\quad
\GeneralOracleInstance(\Witness\concat i)_1=0
\quad\text{for every }i\in\{1,\ldots,m\}.
\end{equation}
Here $\concat$ denotes concatenation, $i$ is written as a binary string of fixed length, and the subscript $1$ denotes the first bit of the oracle's answer.

Think of these bits as a table with $2^m$ rows and $m$ columns (\Cref{fig:table}). Each row is labeled by a candidate witness $\Witness$, and column $i$ contains $\GeneralOracleInstance(\Witness\concat i)_1$. The claim is that the table contains an all-zero row, and the witness is the label of such a row. Given the witness, $C_\star$ checks the claim by reading the $m$ bits in that row. Without the witness, finding an all-zero row is hard: the rows are independent, each is all zero with probability $2^{-m}$, and polynomially many oracle queries have only an exponentially small chance of finding one. 

The table is a simplified version of the drug-discovery example. A row label is the structure of a molecule, a column is a test, and a zero means that the molecule passed that test. The claim is that some molecule passes every test, and a witness is such a molecule. Asking for a test result means submitting the structure, so a verifier cannot ask about a molecule without revealing it. Our lower bound also assumes that the test outcomes are independent random bits, which is what the random oracle provides.

\parhead{The verifier must read the witness row}
Suppose, towards a contradiction, that $(\IARGProver,\IARGVerifier)$ is a witness-hiding relativized argument for $C_\star$. Consider a cheating prover $\Malicious{\IARGProver}$ that chooses a random row label $\Witness$ and runs $\IARGProver$ on $(0^n,\Witness)$, but answers its oracle queries using a modified oracle $\GeneralOracleInstance'$ (\Cref{fig:two-worlds}). This oracle agrees with $\GeneralOracleInstance$ except that the first bits of all answers in row $\Witness$ are set to zero. $\Malicious{\IARGProver}$ can supply these altered answers to its internal copy of $\IARGProver$; the verifier $\IARGVerifier$ still queries the real oracle $\GeneralOracleInstance$.

\begin{figure}[!b]
\centering
\begin{tikzpicture}[
  cell/.style={draw=gray!60, minimum width=7mm, minimum height=5.5mm, inner sep=0pt, anchor=center, font=\small},
  note/.style={font=\small, align=center, inner sep=1pt},
]
\matrix (L) [matrix of nodes, nodes={cell}, column sep=-\pgflinewidth, row sep=-\pgflinewidth,
  row 4/.style={nodes={fill=red!10}}] {
 1 & 0 & $\cdots$ & 1 \\
 0 & 1 & $\cdots$ & 0 \\
 $\vdots$ & $\vdots$ & & $\vdots$ \\
 1 & 0 & $\cdots$ & 1 \\
 $\vdots$ & $\vdots$ & & $\vdots$ \\
 1 & 1 & $\cdots$ & 0 \\
};
\node[left=2mm of L-1-1, font=\small] {$0\cdots00$};
\node[left=2mm of L-2-1, font=\small] {$0\cdots01$};
\node[left=2mm of L-4-1, font=\small] {$\Witness$};
\node[left=2mm of L-6-1, font=\small] {$1\cdots11$};
\node[note, above=2.5mm of L] {real oracle $\GeneralOracleInstance$\\[-1pt] {\footnotesize queried by the verifier}};
\node[note, below=2mm of L] {no all-zero row: the claim is false};
\matrix (R) [matrix of nodes, nodes={cell}, column sep=-\pgflinewidth, row sep=-\pgflinewidth,
  row 4/.style={nodes={fill=green!20}}, right=22mm of L] {
 1 & 0 & $\cdots$ & 1 \\
 0 & 1 & $\cdots$ & 0 \\
 $\vdots$ & $\vdots$ & & $\vdots$ \\
 0 & 0 & $\cdots$ & 0 \\
 $\vdots$ & $\vdots$ & & $\vdots$ \\
 1 & 1 & $\cdots$ & 0 \\
};
\node[note, above=2.5mm of R] {modified oracle $\GeneralOracleInstance'$\\[-1pt] {\footnotesize fed to the internal honest prover}};
\node[note, below=2mm of R] {row $\Witness$ is all zero: the claim is true};
\draw[<->, gray] (L-4-4.east) -- node[above, font=\footnotesize] {differ only here} (R-4-1.west);
\end{tikzpicture}
\caption{The two worlds in the proof. Left: the real oracle $\GeneralOracleInstance$, which the verifier queries; no row is all zero, so the claim is false. Right: the oracle $\GeneralOracleInstance'$ that the cheating prover feeds to its internal copy of the honest prover; row $\Witness$ is overwritten with zeros, so the claim is true and $\Witness$ is a witness. The tables agree outside row $\Witness$, so a verifier that never reads that row receives the same oracle answers, and the same prover messages, in both worlds.}
\label{fig:two-worlds}
\end{figure}

Now suppose the real table has no all-zero row, which happens with constant probability. Then $C_\star^{\GeneralOracleInstance}(0^n,\Witness')=0$ for every $\Witness'$, while $C_\star^{\GeneralOracleInstance'}(0^n,\Witness)=1$. Consider two interactions in which the prover sends exactly the same messages.
\begin{itemize}[noitemsep]
  \item \emph{The honest interaction} between $\IARGProver$ on input $(0^n,\Witness)$ and $\IARGVerifier$ on input $0^n$, both with oracle $\GeneralOracleInstance'$. By completeness, its decision is $1$ with high probability, since $C_\star^{\GeneralOracleInstance'}(0^n,\Witness)=1$.
  \item \emph{The cheating interaction} between $\Malicious{\IARGProver}$ on input $0^n$ and $\IARGVerifier$ on input $0^n$, both with oracle $\GeneralOracleInstance$ ($\Malicious{\IARGProver}$ feeds $\GeneralOracleInstance'$ to its internal copy of $\IARGProver$). By soundness against $\Malicious{\IARGProver}$, its decision is $0$ with high probability, since $C_\star^{\GeneralOracleInstance}(0^n,\Witness')=0$ for every $\Witness'$.
\end{itemize}
The two oracles agree outside row $\Witness$. As long as $\IARGVerifier$ does not query row $\Witness$, it receives the same oracle answers and the same prover messages in both interactions, and therefore reaches the same decision. Since the two decisions must differ, $\IARGVerifier$ queries row $\Witness$ with high probability in the honest interaction with $\GeneralOracleInstance'$. Making that query requires $\IARGVerifier$ to know $\Witness$. So the honest verifier learns a witness that it could not have found on its own, which violates witness hiding, and hence zero knowledge as well (\Cref{lem:zk-implies-wh}). Nothing in the argument required $\IARGVerifier$ to be succinct, which is why the theorem needs no succinctness requirement (\Cref{thm:wh-lower-bound-rom}).

\subsection{Debate}
\label{sec:techniques-debate}

We now show that the lower bound extends to debate. This contrasts with succinctness: a second prover lets debate achieve succinct verification where a single prover cannot \cite{BCIP24}, but it does not make privacy possible. Recall that a \emph{debate} involves a prover $\DebateProver$, an opponent $\DebateOpponent$, and a verifier $\Judge$ \cite{ICA18,BCIP24}. All three know the public input $\Instance$ and can query the oracle $\GeneralOracleInstance$. The prover $\DebateProver$ tries to establish that there exists a witness $\Witness$ such that $D^{\GeneralOracleInstance}(\Instance,\Witness)=1$, while the opponent $\DebateOpponent$ challenges this claim. Completeness requires that, if the claim is true, the honest prover $\DebateProver$ convinces $\Judge$ to accept regardless of the opponent's strategy. Soundness requires that, if the claim is false, the honest opponent $\DebateOpponent$ makes $\Judge$ reject regardless of the cheating prover's strategy.

There are two natural ways to define privacy for debate, depending on which provers receive the secret witness. One option is to give $\Witness$ to both $\DebateProver$ and $\DebateOpponent$ and require that it remain hidden from $\Judge$. This fails in general because either prover may be dishonest, and a dishonest prover can simply send $\Witness$ to $\Judge$. The other option is to give $\Witness$ only to $\DebateProver$, while $\DebateOpponent$ and $\Judge$ receive only $\Instance$ and oracle access. In this case, privacy must hold against $\DebateOpponent$ and $\Judge$ together, since they may cooperate. We require only that $\DebateOpponent$ and $\Judge$, when both follow the protocol, cannot recover $\Witness$ from what they see together. This is the weakest privacy requirement for this option: a malicious $\DebateOpponent$ or $\Judge$ can in particular follow the protocol, so privacy against malicious parties implies it, and zero knowledge, which requires their view to reveal nothing about $\Witness$, in particular requires that it not reveal $\Witness$. Ruling it out therefore rules out every stronger requirement.

For this second definition, our lower bound applies by a reduction to the single-prover case. Let $\IARGVerifier$ be the verifier that runs $\DebateOpponent$ and $\Judge$ internally, exchanges messages with $\DebateProver$ on their behalf, and outputs the decision of $\Judge$. Then $(\DebateProver,\IARGVerifier)$ is a single-prover argument for the same claim:
\begin{itemize}[noitemsep]
  \item it is complete, because the honest $\DebateProver$ convinces $\Judge$ against every opponent, in particular against $\DebateOpponent$;
  \item it is sound, because the soundness of the debate is exactly the guarantee that $\DebateOpponent$ defeats every cheating $\Malicious{\IARGProver}_1$;
  \item it is witness hiding if the debate is, because $\IARGVerifier$ knows exactly what $\DebateOpponent$ and $\Judge$ know together.
\end{itemize}
Hence a witness-hiding debate for the hard computation $C_\star$ would give a witness-hiding argument for it, which \Cref{ithm:wh} rules out. \Cref{cor:wh-debate} states the result, and \Cref{sec:debate} gives the reduction (\Cref{lem:debate-folding}).

\subsection{Construction with signed oracles}
\label{sec:techniques-signed-oracles}

We now show how a signed oracle makes succinct zero-knowledge verification possible for every oracle-aided computation (\Cref{ithm:signing}). Recall from \Cref{sec:techniques-lower-bound} that the lower bound rests on one fact: the only way for the verifier to check that an oracle answer is genuine is to make the query, and for some computations the query requires knowing the witness. A signed oracle $\SignedOracleInstance$ (\Cref{sec:our-results}) removes this obstacle. Every answer comes with a signature on the query and the answer, and anyone holding $\SigVK$ can check the signature without querying the oracle, so certifying an answer no longer requires querying the oracle itself.

Here is a first attempt. The prover runs $D$ on $(\Instance,\Witness)$ with the signed oracle and sends the verifier the entire record of the execution: every query it made, the answer, and the signature. The verifier checks that the record is an accepting execution of $D$ and that every signature in it is valid under $\SigVK$, without querying the oracle. This is sound, but it is not zero knowledge in general: the record contains the prover's queries, which may contain the confidential data (in the computation of \Cref{sec:techniques-lower-bound}, the queries are the witness itself). So the prover must keep $\Witness$ and the record to itself, and instead prove to the verifier that it holds a witness and a record such that (i) the record is an accepting execution of $D$ on $(\Instance,\Witness)$ and (ii) every signature in the record is valid under $\SigVK$. Given $\Witness$ and the record, checking (i) and (ii) is an ordinary computation that never calls $\GeneralOracleInstance$.

Specifically, the prover proves that for $(\Instance,\SigVK)$ there exist $\Witness$ and $((q_i,a_i,\sigma_i))_i$ such that $D$ on $(\Instance,\Witness)$ makes the queries $q_1,q_2,\ldots$, receives the answers $a_1,a_2,\ldots$, and accepts, and $\SigVerify(\SigVK,(q_i,a_i),\sigma_i)=1$ for every $i$. This is an $\NP$ statement that does not depend on any oracle: given $\Witness$ and $((q_i,a_i,\sigma_i))_i$, it can be checked in polynomial time. Thus we can use standard techniques from cryptography to prove it, namely a \emph{zero-knowledge argument of knowledge}. Such an argument lets a prover convince a verifier that it knows a valid witness for an $\NP$ statement while revealing nothing about that witness beyond the truth of the statement. Arguments of this kind with succinct verification exist assuming collision-resistant hash functions \cite{Kilian92,BarakG08}.

We apply this argument to the statement above, with public input $(\Instance,\SigVK)$ and private witness $(\Witness,((q_i,a_i,\sigma_i))_i)$. The resulting protocol has the three properties we want:
\begin{itemize}[noitemsep]
  \item it is sound, because the knowledge requirement lets us extract the record from a convincing prover, and for a false claim the record must contain a fabricated oracle answer with a valid signature, a forgery that the security of the signature scheme rules out;
  \item it is zero knowledge, because the argument hides the original witness, the queries, and the answers;
  \item it has succinct verification, because the verifier never queries the underlying oracle $\GeneralOracleInstance$, and checks a succinct argument.
\end{itemize}
Combining a signed oracle with an ordinary zero-knowledge argument of knowledge therefore gives succinct zero-knowledge verification of the original oracle-aided computation (\Cref{thm:zk-signed-oracle}). Our construction keeps the argument of knowledge simple to demonstrate feasibility; replacing it by a modern hash-based succinct argument makes the protocol concretely efficient (\Cref{rem:efficiency}).

\doclearpage

\section{Preliminaries}
\label{sec:prelim}
Some of our definitions in this section are taken verbatim from \cite{BarbaraCG25}.

For a set $S$, we denote by $2^S$ the set of all subsets of $S$. All sets considered in this paper are finite. For every finite set, we can define a bijective binary encoding. Therefore, without loss of generality, we assume every set considered in this paper is a subset of $\bits^*$.

We write $x\gets \Distribution$ to denote that $x$ is a sample of the distribution $\Distribution$. We write $x\gets \DistDomain$ to denote that $x$ is sampled uniformly at random from the non-empty set $\DistDomain$. We write $x\in\Distribution$ to denote that $x$ is in the support of $\Distribution$, i.e. the value $x$ is sampled with non-zero probability from $\Distribution$.

\subsection{Oracles, languages and relations}
\begin{definition}
For all finite sets $\FunDomain$ and $\FunRange$, the set $\Functions{\FunDomain}{\FunRange}$ is the set of all functions $\GeneralOracleInstance\colon \FunDomain\to\FunRange$. 
\end{definition}

\begin{definition}[Oracle]
\label{def:oracle}
An \defemph{oracle} is a collection $\GeneralOracle=\{\GeneralOracle_{\Security,\InstanceSize}\}_{\Security\in\N,\InstanceSize\in\N}$ where each $\GeneralOracle_{\Security,\InstanceSize}$ is a distribution over functions $\Functions{\FunDomain_{\Security,\InstanceSize}}{\FunRange_{\Security,\InstanceSize}}$ for some finite sets $\FunDomain_{\Security,\InstanceSize},\FunRange_{\Security,\InstanceSize}$.

We write $\GeneralOracleInstance\in\GeneralOracle$ to denote that $\GeneralOracleInstance$ is in the support of $\GeneralOracle_{\Security,\InstanceSize}$ for some $\Security\in\N,\InstanceSize\in\N$. An oracle $\GeneralOracle$ has \defemph{input size bound} $\OracleInputSize$ if for every $\Security,\InstanceSize\in\N$, $\FunDomain_{\Security,\InstanceSize} \subseteq \bits^{\leq\OracleInputSize(\Security,\InstanceSize)}$.

The empty oracle $\EmptyOracle$ is the oracle whose only function $f\in\EmptyOracle$ is the function with empty domain.
\end{definition}

\begin{definition}[Relativized language]
Let $\GeneralOracle$ be an oracle. A relativized language $\Language_{\GeneralOracle}$ is a collection of languages indexed by functions $\GeneralOracleInstance\in \GeneralOracle_{\Security,\InstanceSize}$ for $\Security,\InstanceSize\in\N$, namely, 
\begin{equation*}
  \Language_{\GeneralOracle}\coloneq \{\Language_\GeneralOracleInstance\}_{\Security,\InstanceSize\in\N,\GeneralOracleInstance\in{\GeneralOracle_{\Security,\InstanceSize}}}\enspace.
\end{equation*}
\end{definition}

\begin{definition}[Relativized relation]
\label{def:relativized-relation}
Let $\GeneralOracle$ be an oracle.
A relativized relation $\Relation_{\GeneralOracle}$ is a collection of relations indexed by a security bound $\Security\in\N$ and instance size $\InstanceSize\in\N$ and functions $\GeneralOracleInstance\in \GeneralOracle_{\Security,\InstanceSize}$, namely, 
\begin{equation*}
  \Relation_{\GeneralOracle}\coloneq \{\Relation_\GeneralOracleInstance\}_{\Security,\InstanceSize\in\N,\GeneralOracleInstance\in{\GeneralOracle_{\Security,\InstanceSize}}}\enspace.
\end{equation*}
The relativized language corresponding to the relativized relation $\Relation_\GeneralOracle$, denoted $\Language(\Relation_\GeneralOracle)$, is defined as follows:
\begin{equation*}
  \Language(\Relation_\GeneralOracle)\coloneq \left\{\Language(\Relation_\GeneralOracleInstance)\right\}_{\Security,\InstanceSize\in\N,\GeneralOracleInstance\in{\GeneralOracle_{\Security,\InstanceSize}}}\enspace,
\end{equation*}
where each $\Language(\Relation_\GeneralOracleInstance)\coloneq \left\{\Instance:\exists\,\Witness \text{ s.t. } (\Instance,\Witness)\in \Relation_\GeneralOracleInstance\right\}$. 
\end{definition}

\begin{remark}
Let $\Relation$ be a relation and $\GeneralOracle$ be an oracle. We can view $\Relation$ as a relativized relation $\{\Relation_\GeneralOracleInstance\}_{\GeneralOracleInstance\in{\GeneralOracle}}$ where $\Relation_\GeneralOracleInstance\coloneq\Relation$ for every $\GeneralOracleInstance\in{\GeneralOracle}$.
\end{remark}

\begin{definition}
Let $\Relation_{\GeneralOracle}$ be a relativized relation. For $\GeneralOracleInstance\in\GeneralOracle$ and $\Instance\in\bits^*$, the \defemph{set of witnesses for $\Instance$ in $\Relation_\GeneralOracleInstance$} is $\WitnessesOf{\GeneralOracleInstance}{\Instance}{\Relation_\GeneralOracle} \coloneq \{
  \Witness \in \bits^* \colon (\Instance,\Witness) \in \Relation_\GeneralOracleInstance
  \}$.
\end{definition} 

\begin{definition}
For every family of pairs of sets $((\FunDomain_{\Security,\InstanceSize},\FunRange_{\Security,\InstanceSize}))_{\Security,\InstanceSize\in\N}$, the \defemph{random oracle over $((\FunDomain_{\Security,\InstanceSize},\FunRange_{\Security,\InstanceSize}))_{\Security,\InstanceSize\in\N}$} is the collection $\{\RandomOracle_{\Security,\InstanceSize}\}_{\Security,\InstanceSize\in\N}$, where each $\RandomOracle_{\Security,\InstanceSize}$ is the uniform distribution over $\Functions{\FunDomain_{\Security,\InstanceSize}}{\FunRange_{\Security,\InstanceSize}}$.
\end{definition}

\subsection{Turing machines and complexity classes with oracles}
\newcommand{\QueryState}{\textsc{Query}}
\newcommand{\AnswerState}{\textsc{Answer}}

To define the complexity class $\NTIME^\GeneralOracle[t]$, we consider Turing machines that can query oracles. 
An oracle machine is a machine that has \emph{black-box} access to a function $f:\FunDomain\rightarrow\FunRange$ that the machine can query. Each query costs the machine a single computational step. 

\begin{definition}
\label{def:oracle-machine}
Let $\GeneralOracle$ be an oracle. An \defemph{oracle Turing machine} $M$ is a Turing machine that has two additional special tapes, the \emph{oracle query tape} and the \emph{oracle answer tape}, and two additional special states, the $\QueryState$ state and the $\AnswerState$ state. 
Given input $\Instance\in\bits^*$ and oracle access to a function $\GeneralOracleInstance\in\GeneralOracle_{\Security,\InstanceSize}$ for some $\Security,\InstanceSize$, $M^\GeneralOracleInstance(\Instance)$ works as follows: 
\begin{itemize}[noitemsep]
  \item The input $\Instance$ is written in the input tape.
  \item The execution of $M$ proceeds as a normal Turing machine computation except that, when $M$ intends to make a query to the oracle, then it enters the $\QueryState$ state. 
  \item In the $\QueryState$ state, $M$ writes its query $q\in\bits^*$ on the query tape. 
  \item In the following step, $M$ enters the $\AnswerState$ state and the content in the answer tape becomes $\GeneralOracleInstance(q)$ if $q$ is in the domain of $\GeneralOracleInstance$ and empty otherwise.
\end{itemize} 
We call the function that maps the tuple of the input $x$  and the oracle answers to the first $i-1$ oracle queries to $M$'s $i$-th query the \emph{$i$-th query function of $M$}.
\end{definition}

We extend the definition of oracle Turing machines to deterministic oracle Turing machines, nondeterministic oracle Turing machines, and probabilistic oracle Turing machines with the standard definitions of deterministic Turing machines, nondeterministic Turing machines, and probabilistic Turing machines in prior literature. Throughout this paper, we focus on oracle algorithms, which are implemented as oracle Turing machines.

\begin{definition}
\label{def:view}
The \defemph{view} of a probabilistic interactive oracle algorithm $\GenericOracleAlgorithm$ is a function $\ViewFunction{\GenericOracleAlgorithm}$ that maps the oracle $\GeneralOracleInstance$, the input $\GenericOracleAlgorithmInput$, the messages $(\IARGVerifierMessage_i)_{i\in[k]}$ that the algorithm receives, and the randomness $\GenericOracleAlgorithmRandomness$ to the tuple $(\GenericOracleAlgorithmInput, \GenericOracleAlgorithmRandomness,  (\IARGVerifierMessage_i)_{i\in[k]},\QueryAnswerList\coloneq((\OracleQuery_i,\GeneralOracleInstance(\OracleQuery_i)))_{i\in[\GenericQueryBound]})$, 
where $(\OracleQuery_i)_{i\in[\GenericQueryBound]}$ 
is the list of all oracle queries of $\GenericOracleAlgorithm^\GeneralOracleInstance(\GenericOracleAlgorithmInput),\GenericOracleAlgorithm^\GeneralOracleInstance(\GenericOracleAlgorithmInput,\IARGVerifierMessage_1)
,\dots,$ and $\GenericOracleAlgorithm^\GeneralOracleInstance(\GenericOracleAlgorithmInput,\IARGVerifierMessage_1,\dots,\IARGVerifierMessage_k)$. 

For every oracle $\GeneralOracleInstance$, input $\GenericOracleAlgorithmInput$, messages $(\IARGVerifierMessage_i)_{i\in[k]}$ and randomness $\GenericOracleAlgorithmRandomness$, the output of $\GenericOracleAlgorithm^{\GeneralOracleInstance}(\GenericOracleAlgorithmInput,(\IARGVerifierMessage_i)_{i\in[k]};\GenericOracleAlgorithmRandomness)$ is a function of $\OracleAlgorithmView{\GenericOracleAlgorithm}{\GeneralOracleInstance}{\GenericOracleAlgorithmInput}{(\IARGVerifierMessage_i)_{i\in[k]}}{\GenericOracleAlgorithmRandomness}$.
\end{definition}

\begin{definition}
\label{def:interaction}
Let $\GenericOracleAlgorithm_0$ and $\GenericOracleAlgorithm_1$ be interactive oracle algorithms, interacting for $\IARGRoundComplexity$ rounds, with respective input domains $\GenericOracleAlgorithmInputSet_0$ and $\GenericOracleAlgorithmInputSet_1$ and randomness complexities $\GenericOracleAlgorithmRandomnessComplexity_0$ and $\GenericOracleAlgorithmRandomnessComplexity_1$. For $\GeneralOracleInstance_0,\GeneralOracleInstance_1\in\GeneralOracle$, $\GenericOracleAlgorithmInput_0\in\GenericOracleAlgorithmInputSet_0$, $\GenericOracleAlgorithmInput_1\in\GenericOracleAlgorithmInputSet_1$, we denote the random experiment 
\begin{equation*}
\left\{
\begin{array}{l}
\GenericOracleAlgorithmRandomness_0 \gets \bits^{\GenericOracleAlgorithmRandomnessComplexity_0}\\
\GenericOracleAlgorithmRandomness_1 \gets \bits^{\GenericOracleAlgorithmRandomnessComplexity_1}\\
\text{For }i\in[\IARGRoundComplexity]:\\
\quad \IARGProverMessage_i \coloneq \GenericOracleAlgorithm_0^{\GeneralOracleInstance_0}(\GenericOracleAlgorithmInput_0,(\IARGVerifierMessage_j)_{j< i};\GenericOracleAlgorithmRandomness_0)\\
\quad \IARGVerifierMessage_i \coloneq \GenericOracleAlgorithm_1^{\GeneralOracleInstance_1}(\GenericOracleAlgorithmInput_1,(\IARGProverMessage_j)_{j\leq i};\GenericOracleAlgorithmRandomness_1)\\
\IARGDecision \coloneq \IARGVerifierMessage_{\IARGRoundComplexity}\\
\IARGTranscript \coloneq (\IARGProverMessage_1,\IARGVerifierMessage_1,\dots,\IARGProverMessage_\IARGRoundComplexity,\IARGVerifierMessage_\IARGRoundComplexity)
\end{array}
\right\}
\end{equation*}
by $\IARGDecision \xleftarrow{\GenericOracleAlgorithmRandomness_0,\GenericOracleAlgorithmRandomness_1,\IARGTranscript} \langle \GenericOracleAlgorithm_0^{\GeneralOracleInstance_0}(\GenericOracleAlgorithmInput_0),\GenericOracleAlgorithm_1^{\GeneralOracleInstance_1}(\GenericOracleAlgorithmInput_1)\rangle$. When $\GenericOracleAlgorithmRandomness_0,\GenericOracleAlgorithmRandomness_1,\IARGTranscript$ are irrelevant, we omit them from the notation.

If $\GenericOracleAlgorithmRandomnessComplexity_0=\GenericOracleAlgorithmRandomnessComplexity_1=0$, we write $\IARGDecision \xleftarrow{\IARGTranscript} \langle \GenericOracleAlgorithm_0^{\GeneralOracleInstance_0}(\GenericOracleAlgorithmInput_0),\GenericOracleAlgorithm_1^{\GeneralOracleInstance_1}(\GenericOracleAlgorithmInput_1)\rangle$. 
\end{definition}

\parhead{Nondeterministic time}
A nondeterministic machine $M$ is an \defemph{ND-decider} for a relation $\Relation$ if for every $(\Instance,\Witness)$ it holds that $M(\Instance,\Witness)=1$ if and only if $(\Instance,\Witness)\in \Relation$, where $M(\Instance,\Witness)$ denotes the decision of $M$ on the input $\Instance$ on the branch $\Witness$.

\begin{definition}[Relativized $\NTIME$]
\label{def:relativized-ntime}
Let $\GeneralOracle$ be an oracle. Let $t$ be a time-constructible function. The complexity class $\NTIME^\GeneralOracle[t]$ is the set of relativized relations $\Relation_{\GeneralOracle}=\{\Relation_\GeneralOracleInstance\}_{\Security,\InstanceSize\in\N,\GeneralOracleInstance\in{\GeneralOracle_{\Security,\InstanceSize}}}$ such that there exists an oracle nondeterministic machine $M$ that runs in time $O(t(\InstanceSize))$ and, for every $\Security,\InstanceSize\in\N$,
\begin{equation*}
\prob{
M^\GeneralOracleInstance \text{ is an ND-decider for } \Relation_\GeneralOracleInstance
\;\middle\vert\;
\GeneralOracleInstance\gets\GeneralOracle_{\Security,\InstanceSize}
}=1\enspace.
\end{equation*}
We call $M$ an \defemph{ND-decider} for $\Relation_\GeneralOracle$.
\end{definition}

\begin{definition}[Universal relation \cite{BarakG08}]
\label{def:universal-relation}
The \defemph{universal relation} $\URel$ is the set of pairs $((M,\Instance,t),\Witness)$ such that the Turing machine $M$ accepts $(\Instance,\Witness)$ within $t$ steps and $\abs{\Witness}\leq t$.
\end{definition}

\subsection{Relativized interactive arguments in an oracle model}
\label{sec:general-oracle-relativized-args}

Let $\GeneralOracle$ be an oracle. A \textbf{relativized interactive argument} in the oracle model $\GeneralOracle$ for a relativized relation $\Relation_\GeneralOracle$ is a pair of interactive oracle algorithms $\IARG=(\IARGProver,\IARGVerifier)$ that satisfies the following properties. Both algorithms have access to the same oracle function $\GeneralOracleInstance\in\GeneralOracle_{\Security,\InstanceSize}$. They interact as in \cref{def:interaction}. Our arguments have no trusted setup: public parameters, such as those of a commitment scheme, are sampled by the verifier and sent as its first message, and the prover aborts if they are not well formed, so zero knowledge has to hold for every well-formed choice of the parameters.

\begin{definition}[Completeness]
\label{def:rel-arg-adaptive-completeness-gom}
$\IARG=(\IARGProver,\IARGVerifier)$ for a relativized relation $\Relation_\GeneralOracle$ has \defemph{completeness error $\IARGCompleteness$} if for every security parameter $\Security\in\N$, instance size $\InstanceSize\in\N$, instance $\Instance\in\bits^\InstanceSize$, and adversary $\IARGCompletenessAdv$,
\begin{equation*}
\prob{
\begin{array}{l}
  \IARGDecision=0 \\
  \land\, (\Instance,\Witness)\in\Relation_\GeneralOracleInstance\\
\end{array}
\;\middle\vert\;
\begin{array}{l}
  \GeneralOracleInstance\gets {\GeneralOracle_{\Security,\InstanceSize}}\\
  \Witness\gets \IARGCompletenessAdv^\GeneralOracleInstance(\Instance)\\
  \IARGDecision\gets \langle \IARGProver^\GeneralOracleInstance(1^\Security,(\Instance,\Witness)), \IARGVerifier^\GeneralOracleInstance(1^\Security,\Instance)\rangle
\end{array}
}
\leq \IARGCompleteness(\Security,\InstanceSize)
\enspace.
\end{equation*}
\end{definition}

\begin{definition}[Soundness]
\label{def:rel-arg-adaptive-soundness-gom}
$\IARG=(\IARGProver,\IARGVerifier)$ for a relativized relation $\Relation_\GeneralOracle$ has \defemph{soundness error $\IARGSoundness$} if for every security parameter $\Security\in\N$, instance size bound $\InstanceSize\in\N$, instance $\Instance\in\bits^\InstanceSize$, adversary query bound $\IARGSoundnessAdvQueryBound\in\N$, adversary time bound $\IARGSoundnessAdvTimeBound\in\N$, and $\IARGSoundnessAdvQueryBound$-query $\IARGSoundnessAdvTimeBound$-time adversary $\Malicious{\IARGProver}$,
\begin{equation*}
\prob{
\begin{array}{l}
  \IARGDecision=1\\
  \land\,\Instance\notin \Language(\Relation_\GeneralOracleInstance)
\end{array}
\;\middle\vert\;
\begin{array}{l}
  \GeneralOracleInstance\gets {\GeneralOracle_{\Security,\InstanceSize}}\\
  \IARGDecision\gets \langle \Malicious{\IARGProver}^\GeneralOracleInstance(\Instance), \IARGVerifier^\GeneralOracleInstance(1^\Security,\Instance)\rangle
\end{array}
}\leq \IARGSoundness(\Security,\InstanceSize,\IARGSoundnessAdvQueryBound,\IARGSoundnessAdvTimeBound)\enspace.
\end{equation*}
\end{definition}

\parhead{Efficiency measures}
We consider the following efficiency measures of $\IARG$, each a function of the security parameter $\Security$ and the instance size $\InstanceSize$: the round complexity $\IARGRoundComplexity$; the prover communication complexity $\IARGProverComm$, in bits; the query complexity $\IARGVerifierQuery$ and the running time $\IARGVerifierTime$ of the verifier; the query complexity $\IARGProverQuery$ and the running time $\IARGProverTime$ of the honest prover; and the randomness complexities $\IARGVerifierRandomnessComplexity$ and $\IARGProverRandomnessComplexity$ of the verifier and the honest prover.

\begin{definition}
\label{def:wh}
Let $\IARG=(\IARGProver,\IARGVerifier)$ be a relativized interactive argument for a relativized relation $\Relation_\GeneralOracle$. $\IARG$ has \defemph{witness-hiding error $\IARGWitnessHiding$} if for every security parameter $\Security\in\N$, instance size bound $\InstanceSize\in\N$, instance $\Instance\in\bits^\InstanceSize$, adversary $\WHSampleAdversary$, and $\WHAdversaryQuery(\Security,\InstanceSize)$-query $\WHAdversaryTime(\Security,\InstanceSize)$-time adversary $\WHOutAdversary$,
\begin{equation*}
\prob{
\begin{array}{l}
(\Instance,\Witness')\in\Relation_\GeneralOracleInstance
\,\land\, (\Instance,\Witness)\in\Relation_\GeneralOracleInstance
\end{array}
\;\middle\vert\;
\begin{array}{l}
  \GeneralOracleInstance\gets {\GeneralOracle_{\Security,\InstanceSize}} \\
  \Witness \gets \WHSampleAdversary^\GeneralOracleInstance(\Instance) \\
  \IARGDecision \xleftarrow{\IARGProverRandomness,\IARGVerifierRandomness,\IARGTranscript} \langle \IARGProver^{\GeneralOracleInstance}(1^\Security,(\Instance,\Witness)), \IARGVerifier^{\GeneralOracleInstance}(1^\Security,\Instance) \rangle\\
  \Witness' \gets \WHOutAdversary^{\GeneralOracleInstance}(\Instance,\IARGVerifierRandomness,\IARGTranscript)
\end{array}
}
\leq 
\IARGWitnessHiding(\Security, \InstanceSize, \WHAdversaryQuery,\WHAdversaryTime)
\enspace.
\end{equation*}
\end{definition}

\begin{definition}[Black-box rewinding access]
\label{def:black-box-rewinding}
For a probabilistic interactive oracle algorithm $\mathcal{A}$ and a fixed oracle function $\GeneralOracleInstance$, we write $\mathcal{B}^{\GeneralOracleInstance,\mathcal{A}}$ to denote that $\mathcal{B}$ has oracle access to $\GeneralOracleInstance$ and \defemph{black-box rewinding access} to $\mathcal{A}$ relative to $\GeneralOracleInstance$: $\mathcal{B}$ can start executions of $\mathcal{A}$, each with an independently sampled random tape, send messages to $\mathcal{A}$ to obtain its next message, rewind $\mathcal{A}$ to an earlier point of an execution (with the same random tape), and obtain the view of $\mathcal{A}$ in a completed execution (\cref{def:view}). The oracle queries of $\mathcal{A}$ are answered by $\GeneralOracleInstance$, and they count toward the running time and query complexity of $\mathcal{B}$.
\end{definition}

\begin{definition}[Knowledge soundness]
\label{def:rel-arg-adaptive-knowledge-soundness-gom}
$\IARG=(\IARGProver,\IARGVerifier)$ for a relativized relation $\Relation_\GeneralOracle$ has \defemph{knowledge soundness error $\IARGKnowledgeSoundness$ with extraction time $\KSExtractorTime$ and query complexity $\KSExtractorQuery$} if there exists a probabilistic algorithm $\KSExtractor$ such that for every security parameter $\Security\in\N$, instance size bound $\InstanceSize\in\N$, oracle function $\GeneralOracleInstance\in\GeneralOracle_{\Security,\InstanceSize}$, instance $\Instance\in\bits^\InstanceSize$, adversary query bound $\KSAdversaryQuery\in\N$, adversary time bound $\KSAdversaryTime\in\N$, and $\KSAdversaryQuery$-query $\KSAdversaryTime$-time malicious prover $\Malicious{\IARGProver}$,
\begin{equation*}
\prob{
\begin{array}{l}
  (\Instance,\Witness)\notin\Relation_\GeneralOracleInstance\\
  \land\, \IARGDecision=1
\end{array}
\;\middle\vert\;
\begin{array}{l}
  \IARGDecision \xleftarrow{\IARGTranscript} \langle \Malicious{\IARGProver}^\GeneralOracleInstance(\Instance), \IARGVerifier^\GeneralOracleInstance(1^\Security,\Instance)\rangle\\
  \Witness \gets \KSExtractor^{\GeneralOracleInstance,\Malicious{\IARGProver}(\Instance)}(1^\Security,\Instance,\IARGTranscript,\IARGDecision)
\end{array}
}
\leq
\IARGKnowledgeSoundness_{\GeneralOracleInstance}(\Security,\InstanceSize,\KSAdversaryQuery,\KSAdversaryTime)
\enspace;
\end{equation*}
moreover, $\KSExtractor$ runs in time $\KSExtractorTime(\Security,\InstanceSize,\KSAdversaryQuery,\KSAdversaryTime)$ and makes at most $\KSExtractorQuery(\Security,\InstanceSize,\KSAdversaryQuery,\KSAdversaryTime)$ queries to the oracle. Above, $\KSExtractor^{\GeneralOracleInstance,\Malicious{\IARGProver}(\Instance)}$ has oracle access to $\GeneralOracleInstance$ and black-box rewinding access to $\Malicious{\IARGProver}(\Instance)$ relative to $\GeneralOracleInstance$ (\cref{def:black-box-rewinding}).
\end{definition}

\begin{definition}
\label{def:mvzk}
A relativized argument $(\IARGProver,\IARGVerifier)$ for a relativized relation $\Relation_\GeneralOracle$ has \defemph{malicious-verifier zero-knowledge error $\IARGZeroKnowledge$ with simulator time $\ZKSimTime$ and query complexity $\ZKSimQuery$} if there exists an $\ZKSimTime$-time $\ZKSimQuery$-query simulator $\ZKSim$ with black-box rewinding access to the malicious verifier as in \cref{def:black-box-rewinding} such that for every security parameter $\Security\in\N$, instance size bound $\InstanceSize\in\N$, oracle function $\GeneralOracleInstance\in\GeneralOracle_{\Security,\InstanceSize}$, instance-witness pair $(\Instance,\Witness)\in\Relation_\GeneralOracleInstance$, auxiliary input $\aux\in\bits^*$, which may depend on $\GeneralOracleInstance$, verifier query bound $\ZKAdversaryQuery\in\N$, verifier time bound $\ZKAdversaryTime\in\N$, and $\ZKAdversaryQuery$-query $\ZKAdversaryTime$-time malicious verifier $\ZKAdversary$, the following two distributions are $\IARGZeroKnowledge(\Security,\InstanceSize,\ZKAdversaryQuery,\ZKAdversaryTime)$-close in statistical distance:
\begin{itemize}[noitemsep]
  \item the \emph{simulated view} $\ZKSim^{\GeneralOracleInstance,\ZKAdversary(\Instance,\aux)}(1^\Security,\Instance)$;
  \item the \emph{real view} $\OracleAlgorithmView{\ZKAdversary}{\GeneralOracleInstance}{(\Instance,\aux)}{(\IARGProverMessage_i)_{i\in[\IARGRoundComplexity]}}{\ZKAdversaryRandomness}$ of $\ZKAdversary$ (\cref{def:view}), where $\ZKAdversaryRandomness$ is the randomness of $\ZKAdversary$ and $(\IARGProverMessage_i)_{i\in[\IARGRoundComplexity]}$ are the messages of $\IARGProver$ in the interaction $\langle\IARGProver^\GeneralOracleInstance(1^\Security,(\Instance,\Witness)),\ZKAdversary^\GeneralOracleInstance(\Instance,\aux)\rangle$.
\end{itemize}
\end{definition}

Zero knowledge implies witness hiding whenever witnesses are hard to find from the instance alone.

\begin{lemma}
\label{lem:zk-implies-wh}
Let $\IARG=(\IARGProver,\IARGVerifier)$ be a relativized interactive argument for a relativized relation $\Relation_\GeneralOracle$ with malicious-verifier zero-knowledge error $\IARGZeroKnowledge$, simulator time $\ZKSimTime$, and query complexity $\ZKSimQuery$ (\cref{def:mvzk}). Suppose that for every security parameter $\Security\in\N$, instance size bound $\InstanceSize\in\N$, instance $\Instance\in\bits^\InstanceSize$, and $\GenericQueryBound$-query $\GenericTimeBound$-time algorithm $\mathcal{B}$,
\begin{equation*}
\prob{
(\Instance,\Witness')\in\Relation_\GeneralOracleInstance
\;\middle\vert\;
\begin{array}{l}
  \GeneralOracleInstance\gets {\GeneralOracle_{\Security,\InstanceSize}} \\
  \Witness' \gets \mathcal{B}^{\GeneralOracleInstance}(\Instance)
\end{array}
}
\leq
\WitnessFindingError(\Security,\InstanceSize,\GenericQueryBound,\GenericTimeBound)
\enspace.
\end{equation*}
Then $\IARG$ has witness-hiding error $\IARGWitnessHiding$ (\cref{def:wh}) such that
\begin{equation*}
\IARGWitnessHiding(\Security,\InstanceSize,\WHAdversaryQuery,\WHAdversaryTime)
\leq
\IARGZeroKnowledge(\Security,\InstanceSize,\IARGVerifierQuery,\IARGVerifierTime)
+
\WitnessFindingError\big(\Security,\InstanceSize,\ZKSimQuery(\Security,\InstanceSize,\IARGVerifierQuery,\IARGVerifierTime)+\WHAdversaryQuery,O(\ZKSimTime(\Security,\InstanceSize,\IARGVerifierQuery,\IARGVerifierTime)+\WHAdversaryTime)\big)
\enspace.
\end{equation*}
\end{lemma}
\begin{proof}
Fix $\Security,\InstanceSize\in\N$, an instance $\Instance\in\bits^\InstanceSize$, an adversary $\WHSampleAdversary$, and a $\WHAdversaryQuery$-query $\WHAdversaryTime$-time adversary $\WHOutAdversary$. Let $\ZKSim$ be the simulator of \cref{def:mvzk}, and let $\mathcal{B}$ be the following algorithm. Given $\Instance$ and oracle access to $\GeneralOracleInstance$, it runs $\ZKSim^{\GeneralOracleInstance,\IARGVerifier}(1^\Security,\Instance)$ for the honest verifier $\IARGVerifier$ with an empty auxiliary input, answering the oracle queries of $\IARGVerifier$ with $\GeneralOracleInstance$. It then computes the verifier randomness $\IARGVerifierRandomness$ and the transcript $\IARGTranscript$ from the simulated view, which determines both, and outputs $\WHOutAdversary^{\GeneralOracleInstance}(\Instance,\IARGVerifierRandomness,\IARGTranscript)$. The algorithm $\mathcal{B}$ makes at most $\ZKSimQuery+\WHAdversaryQuery$ queries and runs in time $O(\ZKSimTime+\WHAdversaryTime)$.

Fix $\GeneralOracleInstance\in\GeneralOracle_{\Security,\InstanceSize}$ and $\Witness$ such that $(\Instance,\Witness)\in\Relation_\GeneralOracleInstance$. By \cref{def:mvzk}, the simulated view is $\IARGZeroKnowledge$-close in statistical distance to the view of $\IARGVerifier$ in its interaction with $\IARGProver^{\GeneralOracleInstance}(1^\Security,(\Instance,\Witness))$. Statistical distance does not increase when we apply $\WHOutAdversary^{\GeneralOracleInstance}$ and test membership in $\Relation_\GeneralOracleInstance$. Hence the probability that $\WHOutAdversary$ outputs a valid witness after the real interaction is at most $\IARGZeroKnowledge$ plus the probability that $\mathcal{B}^{\GeneralOracleInstance}(\Instance)$ outputs a valid witness. Averaging over $\GeneralOracleInstance\gets\GeneralOracle_{\Security,\InstanceSize}$ and $\Witness\gets\WHSampleAdversary^{\GeneralOracleInstance}(\Instance)$ bounds the probability in \cref{def:wh} by $\IARGZeroKnowledge+\WitnessFindingError$ at the stated resources.
\end{proof}

\begin{definition}[Relativized $\IARG$ in an oracle model]
\label{def:rel-iarg-gom}
Let $\GeneralOracle$ be an oracle. The complexity class
\begin{equation*}
\IARG^{\GeneralOracle}
\left[
\begin{array}{l l l l l l  }
  \emph{\textrm{completeness error}}	& 	\IARGCompleteness & = &\IARGCompleteness(\Security,n)\\
  \emph{\textrm{soundness error}}	&	\IARGSoundness & = & \IARGSoundness(\Security,n,\IARGSoundnessAdvQueryBound,\IARGSoundnessAdvTimeBound)\\
  \emph{\textrm{knowledge soundness error}}	&	\IARGKnowledgeSoundness & = & \IARGKnowledgeSoundness(\Security,n,\KSAdversaryQuery,\KSAdversaryTime)\\
  \emph{\textrm{witness-hiding error}}	&	\IARGWitnessHiding & = & \IARGWitnessHiding(\Security,n,\WHAdversaryQuery,\WHAdversaryTime)\\
  \emph{\textrm{zero-knowledge error}}	&	\IARGZeroKnowledge & = & \IARGZeroKnowledge(\Security,n,\ZKAdversaryQuery,\ZKAdversaryTime)\\
  \emph{\textrm{verifier query bound}} & \IARGVerifierQuery & = & \IARGVerifierQuery(\Security,n)\\
  \emph{\textrm{verifier running time}}	&	\IARGVerifierTime & = & \IARGVerifierTime(\Security,n)\\
  \emph{\textrm{prover communication}	}&	\IARGProverComm & = &\IARGProverComm(\Security,n)\\
  \emph{\textrm{honest prover query bound}}	&	\IARGProverQuery & = & \IARGProverQuery(\Security,n)\\
  \emph{\textrm{honest prover time bound}}	&	\IARGProverTime & = & \IARGProverTime(\Security,n)\\
\end{array}
\right]
\end{equation*}
is the set of relativized relations $\Relation_{\GeneralOracle}=\{\Relation_\GeneralOracleInstance\}_{\Security,\InstanceSize\in\N, \GeneralOracleInstance\in\GeneralOracle_{\Security,\InstanceSize}}$ that admit a relativized interactive argument $(\IARGProver,\IARGVerifier)$ with the listed errors and efficiency measures. A parameter that is omitted is unconstrained.
\end{definition}

\parhead{Arguments for non-relativized relations}
A relation $\Relation$ that does not depend on the oracle is the relativized relation with $\Relation_\GeneralOracleInstance=\Relation$ for every $\GeneralOracleInstance\in\GeneralOracle$. When, in addition, the honest prover and verifier make no oracle queries, we call $\IARG$ an \defemph{interactive argument for $\Relation$} and say that its knowledge soundness error and zero-knowledge error hold \defemph{relative to $\GeneralOracle$}: the oracle is available only to the adversary, the extractor, and the simulator.


\subsection{Collision-resistant hash functions}
\label{sec:crh}

A keyed hash family is defined as follows.
\begin{definition}
\label{def:crh}
A \defemph{keyed hash family} is a pair of polynomial-time algorithms $\HashFamily=(\HashGen,\HashEval)$, where $\HashGen(1^\Security)$ outputs a key $\HashKey$ and $\HashEval_{\HashKey}\colon\bits^*\to\bits^\Security$ is a deterministic function. It has \defemph{collision-resistance error $\HashCollision$ relative to $\GeneralOracle$} if for every $\Security,\InstanceSize\in\N$, $\GeneralOracleInstance\in\GeneralOracle_{\Security,\InstanceSize}$, and $\GenericTimeBound$-time adversary $\mathcal{A}^{\GeneralOracleInstance}$ making at most $\GenericQueryBound$ queries to $\GeneralOracleInstance$,
\begin{equation*}
\prob{
  \HashEval_{\HashKey}(m)=\HashEval_{\HashKey}(m')
  \;\land\;
  m\neq m'
  \;\middle\vert\;
  \begin{array}{l}
    \HashKey\gets\HashGen(1^\Security)\\
    (m,m')\gets\mathcal{A}^{\GeneralOracleInstance}(1^\Security,\HashKey)
  \end{array}
}
\leq
\HashCollision^{\GeneralOracleInstance}(\Security,\GenericQueryBound,\GenericTimeBound)
\enspace.
\end{equation*}
The family is \defemph{collision resistant relative to $\GeneralOracle$} if for every $\GeneralOracleInstance\in\GeneralOracle_{\Security,\InstanceSize}$, $\HashCollision^{\GeneralOracleInstance}(\Security,\GenericQueryBound,\GenericTimeBound)$ is negligible whenever $\InstanceSize,\GenericQueryBound,\GenericTimeBound$ are polynomially bounded in $\Security$.
\end{definition}

\subsection{Digital signatures}
\label{sec:digital-signatures}

A digital signature scheme is defined as follows.
\begin{definition}
\label{def:signature}
A \defemph{digital signature scheme} for messages in $\bits^*$ is a tuple of polynomial-time algorithms $\SigScheme=(\SigGen,\SigSign,\SigVerify)$ with the following syntax.
\begin{itemize}[nolistsep]
  \item $\SigGen(1^\Security)$ outputs a signing key $\SigSK$ and a verification key $\SigVK$.
  \item $\SigSign(\SigSK,\SigMessage)$ outputs a signature $\Sig$ on a message $\SigMessage$.
  \item $\SigVerify(\SigVK,\SigMessage,\Sig)$ deterministically outputs a bit indicating whether $\Sig$ is a valid signature on $\SigMessage$ under $\SigVK$.
\end{itemize}

The scheme $\SigScheme$ is a digital signature scheme with unforgeability error $\SigUnforgeability$ relative to an oracle $\GeneralOracle$ if the following hold.
\begin{itemize}[noitemsep]
  \item \defemph{Perfect correctness.} For every $\Security\in\N$, every $(\SigSK,\SigVK)\in\SupportOf{\SigGen(1^\Security)}$, and every message $\SigMessage$,
  \begin{equation*}
  \prob{
    \SigVerify(\SigVK,\SigMessage,\Sig)=1
    \;\middle\vert\;
    \Sig\gets\SigSign(\SigSK,\SigMessage)
  }=1
  \enspace.
  \end{equation*}
  \item \defemph{Existential unforgeability under adaptive chosen-message attacks (EUF-CMA) error $\SigUnforgeability$ relative to $\GeneralOracle$.} For every $\Security,\InstanceSize\in\N$, $\GeneralOracleInstance\in\GeneralOracle_{\Security,\InstanceSize}$, and $\GenericTimeBound$-time adversary $\SigAdv^{\GeneralOracleInstance}$ making at most $\GenericQueryBound$ queries in total to $\GeneralOracleInstance$ and to the adaptive signing oracle $\SigSign(\SigSK,\cdot)$,
  \begin{equation*}
\prob{
  \SigVerify(\SigVK,\SigMessage^\star,\Sig^\star)=1
  \;\land\;
  \SigMessage^\star\notin\QuerySet
  \;\middle\vert\;
  \begin{array}{l}
    (\SigSK,\SigVK)\gets\SigGen(1^\Security)\\
    (\SigMessage^\star,\Sig^\star)\gets
      \SigAdv^{\GeneralOracleInstance,\SigSign(\SigSK,\cdot)}(1^\Security,\SigVK)
  \end{array}
}
\leq
\SigUnforgeability^{\GeneralOracleInstance}(\Security,\GenericQueryBound,\GenericTimeBound)
\enspace,
\end{equation*}
where $\QuerySet$ is the set of messages submitted to the signing oracle $\SigSign(\SigSK,\cdot)$. The scheme is \defemph{EUF-CMA secure relative to $\GeneralOracle$} if for every $\GeneralOracleInstance\in\GeneralOracle_{\Security,\InstanceSize}$, $\SigUnforgeability^{\GeneralOracleInstance}(\Security,\GenericQueryBound,\GenericTimeBound)$ is negligible whenever $\InstanceSize,\GenericQueryBound,\GenericTimeBound$ are polynomially bounded in $\Security$.
\end{itemize}
\end{definition}

\begin{theorem}[\cite{NaorY89,Rompel90,KatzK05}]
Assuming the existence of one-way functions (OWF), there exists an EUF-CMA secure digital signature scheme (relative to the empty oracle).
\end{theorem}

Collision-resistant hash functions imply one-way functions, and both constructions are black-box, so the theorem extends directly to any oracle.

\begin{corollary}
\label{cor:digital-signature-exists}
Let $\GeneralOracle$ be an oracle. If there exists a keyed hash family that is collision resistant relative to $\GeneralOracle$ (\cref{def:crh}), then there exists a digital signature scheme that is EUF-CMA secure relative to $\GeneralOracle$.
\end{corollary}

\subsection{Probabilistically checkable proofs}
\label{sec:pcp}

A \textbf{probabilistically checkable proof} (PCP) for a relation $\Relation$ is a pair of probabilistic algorithms $\PCP=(\PCPProver,\PCPVerifier)$ that satisfies the following properties. The prover $\PCPProver(\Instance,\Witness)$ outputs a proof string $\PCPProof\in\bits^{\PCPProofLength}$. The verifier $\PCPVerifier^{\PCPProof}(\Instance;\PCPRandomness)$ receives the instance $\Instance$ and randomness $\PCPRandomness\in\bits^{\PCPRandomnessComplexity}$, has oracle access to $\PCPProof$, and outputs a decision bit; each query is an index $i\in[\PCPProofLength]$ answered by the bit $\PCPProof[i]$. The \defemph{view} of the verifier in the execution $\PCPVerifier^{\PCPProof}(\Instance;\PCPRandomness)$ is $(\PCPRandomness,\PCPQueryIndices,\PCPProof|_{\PCPQueryIndices})$, where $\PCPQueryIndices\subseteq[\PCPProofLength]$ is the set of indices that it queries. The PCP has proof length $\PCPProofLength$, query complexity $\PCPQueryComplexity$, randomness complexity $\PCPRandomnessComplexity$ in bits, and verifier running time $\PCPVerifierTime$; these measures may depend on the instance $\Instance$, for example on $\abs{\Instance}$.

\begin{definition}[Completeness]
\label{def:PCP-completeness}
$\PCP=(\PCPProver,\PCPVerifier)$ for a relation $\Relation$ has \defemph{completeness error $\PCPCompletenessError$} if, for every instance size bound $\InstanceSize\in\N$ and every instance-witness pair $(\Instance,\Witness)\in\Relation$ with $\abs{\Instance}\leq\InstanceSize$,
\begin{equation*}
\prob{\PCPVerifier^\PCPProof(\Instance;\PCPRandomness)=1
\;\middle\vert\;
\begin{array}{l}
\PCPProof \gets \PCPProver(\Instance,\Witness)\\
\PCPRandomness \gets \bits^{\PCPRandomnessComplexity}
\end{array}
}
\geq 1-\PCPCompletenessError(\InstanceSize)
\enspace.
\end{equation*}
It has \defemph{perfect completeness} if $\PCPCompletenessError$ is zero.
\end{definition}

\begin{definition}[Knowledge soundness]
\label{def:pcp-knowledge-soundness}
$\PCP=(\PCPProver,\PCPVerifier)$ for a relation $\Relation$ has \defemph{knowledge soundness error $\PCPKnowledgeSoundness$ with extraction time $\PCPExtractorTime$} if there exists a probabilistic algorithm $\PCPExtractor$ such that, for every instance size bound $\InstanceSize\in\N$, instance $\Instance$ with $\abs{\Instance}\leq\InstanceSize$, and proof string $\PCPProof\in\bits^{\PCPProofLength}$,
\begin{equation*}
\prob{
  \begin{array}{l}
    (\Instance,\Witness) \notin \Relation \\
    \land\, \PCPVerifier^\PCPProof(\Instance;\PCPRandomness)=1
  \end{array}
  \;\middle\vert\;
 \begin{array}{l}
    \PCPRandomness \gets \bits^{\PCPRandomnessComplexity}\\
    \Witness\gets\PCPExtractor(\Instance,\PCPProof)
 \end{array} 
}
\leq 
\PCPKnowledgeSoundness(\InstanceSize)
\enspace;
\end{equation*}
moreover, $\PCPExtractor$ runs in time $\PCPExtractorTime(\InstanceSize)$.
\end{definition}

\begin{definition}[Perfect honest-verifier zero knowledge]
\label{def:hvzk-PCP}
A PCP $\PCP=(\PCPProver,\PCPVerifier)$ is \defemph{perfect honest-verifier zero knowledge} (HVZK) if there exists a probabilistic algorithm $\PCPSimulator$ such that for every $(\Instance,\Witness)\in\Relation$, the following two random variables are identically distributed:
\begin{equation}
\label{eq:PCP-hvzk}
\PCPSimulator(\Instance)
\text{ and }
\left\{(\PCPRandomness,\PCPQueryIndices,\PCPProof|_{\PCPQueryIndices})
\;\middle\vert\;
\begin{array}{l}
\PCPProof\gets\PCPProver(\Instance,\Witness)\\
\PCPRandomness\gets\bits^{\PCPRandomnessComplexity}\\
\PCPQueryIndices \coloneq \text{the indices queried by } \PCPVerifier^\PCPProof(\Instance;\PCPRandomness)
\end{array}
\right\}.
\end{equation}
Moreover, $\PCPSimulator$ runs in time at most $\poly(\abs{\Instance},\PCPVerifierTime)$.
\end{definition}

\begin{theorem}
\label{thm:hvzk-PCP-knowledge}
There is a perfect honest-verifier zero-knowledge PCP (\cref{def:hvzk-PCP}) for the universal relation $\URel$ (\cref{def:universal-relation}) with perfect completeness, knowledge soundness error at most $1/2$, constant query complexity, proof length $\poly(t)$, randomness complexity $O(\log t)$, and verifier running time $\poly(\abs{M}+\abs{\Instance}+\log t)$. Its prover and extractor run in time $\poly(\abs{M}+\abs{\Instance}+t)$.
\end{theorem}
\begin{proof}
Apply the honest-verifier zero-knowledge transformation of \cite[Theorem~5]{IshaiWeiss14}, whose composition step preserves knowledge soundness \cite[Lemma~A.5]{ChiesaFenzi24}, to a constant number of parallel repetitions of the PCP of knowledge for $\URel$ of \cite[Section~3.4]{BCCT11} (obtained from \cite{BGHSV05} via \cite{Valiant08}); sequential repetition on independent proofs preserves honest-verifier zero knowledge and reduces the knowledge soundness error.
\end{proof}

\subsection{Vector commitment schemes}
\label{sec:vc}

A \textbf{vector commitment scheme} is a tuple of polynomial-time algorithms $$\VC=(\VCGen,\VCCommit,\VCOpen,\VCCheck)$$ with the following syntax.
\begin{itemize}[nolistsep]
  \item $\VCGen(1^\Security,\VCLength)$, on input a security parameter $\Security\in\N$ and a message length $\VCLength\in\N$, samples public parameters $\VCParams$.
  \item $\VCCommit(\VCParams,\VCVector)$, on input public parameters $\VCParams$ and a message $\VCVector\in\bits^{\VCLength}$, outputs a commitment $\VCCommitment$ and an auxiliary state $\VCState$.
  \item $\VCOpen(\VCParams,\VCState,Q)$, on input public parameters $\VCParams$, an auxiliary state $\VCState$, and a query set $Q\subseteq[\VCLength]$, outputs an opening proof $\VCOpening$ for $\VCVector[Q]$.
  \item $\VCCheck(\VCParams,\VCCommitment,Q,\VCAnswers,\VCOpening)$, on input public parameters $\VCParams$, a commitment $\VCCommitment$, a query set $Q\subseteq[\VCLength]$, an answer string $\VCAnswers\in\bits^{Q}$, and an opening proof $\VCOpening$, outputs a bit indicating whether $\VCOpening$ is a valid proof that $\VCAnswers$ is the restriction to $Q$ of the message committed in $\VCCommitment$.
\end{itemize}
Since in our protocols the verifier samples the public parameters (\cref{sec:general-oracle-relativized-args}), we assume that membership in the support of $\VCGen(1^\Security,\VCLength)$ can be decided in time $\poly(\Security,\VCLength)$.

\begin{definition}[Completeness]
\label{def:vc-completeness}
$\VC=(\VCGen,\VCCommit,\VCOpen,\VCCheck)$ has \defemph{perfect completeness} if for every security parameter $\Security\in\N$, message length $\VCLength\in\N$, message $\VCVector\in\bits^{\VCLength}$, and query set $Q\subseteq[\VCLength]$,
\begin{equation*}
\prob{
  \VCCheck(\VCParams,\VCCommitment,Q,\VCVector[Q],\VCOpening)=1
  \;\middle\vert\;
  \begin{array}{l}
    \VCParams\gets\VCGen(1^\Security,\VCLength)\\
    (\VCCommitment,\VCState)\gets\VCCommit(\VCParams,\VCVector)\\
    \VCOpening\gets\VCOpen(\VCParams,\VCState,Q)
  \end{array}
}=1
\enspace.
\end{equation*}
\end{definition}

\begin{definition}[Position binding]
\label{def:vc-position-binding}
$\VC=(\VCGen,\VCCommit,\VCOpen,\VCCheck)$ has \defemph{position binding error $\VCBinding$ relative to $\GeneralOracle$} if for every security parameter $\Security\in\N$, instance size bound $\InstanceSize\in\N$, oracle function $\GeneralOracleInstance\in\GeneralOracle_{\Security,\InstanceSize}$, message length $\VCLength\in\N$, query set size $s\in\N$ with $s\leq\VCLength$, and $\GenericQueryBound$-query $\GenericTimeBound$-time adversary $\VCAdv$,
\begin{equation*}
\begin{aligned}
&\prob{
  \begin{array}{l}
    \abs{Q}=\abs{Q'}=s\\
    \land\,\exists\, i\in Q\cap Q'\colon \VCAnswers[i]\neq\VCAnswers'[i]\\
    \land\,\VCCheck(\VCParams,\VCCommitment,Q,\VCAnswers,\VCOpening)=1\\
    \land\,\VCCheck(\VCParams,\VCCommitment,Q',\VCAnswers',\VCOpening')=1
  \end{array}
  \;\middle\vert\;
  \begin{array}{l}
    \VCParams\gets\VCGen(1^\Security,\VCLength)\\
    (\VCCommitment,Q,Q',\VCAnswers,\VCAnswers',\VCOpening,\VCOpening')\gets\VCAdv^{\GeneralOracleInstance}(1^\Security,\VCParams)
  \end{array}
}\\
&\quad\leq
\VCBinding^{\GeneralOracleInstance}(\Security,\VCLength,s,\GenericQueryBound,\GenericTimeBound)
\enspace.
\end{aligned}
\end{equation*}
It is \defemph{position binding relative to $\GeneralOracle$} if for every $\GeneralOracleInstance\in\GeneralOracle_{\Security,\InstanceSize}$, $\VCBinding^{\GeneralOracleInstance}(\Security,\VCLength,s,\GenericQueryBound,\GenericTimeBound)$ is negligible whenever $\InstanceSize,\VCLength,s,\GenericQueryBound,\GenericTimeBound$ are polynomially bounded in $\Security$.
\end{definition}

\begin{definition}[Hiding]
\label{def:vc-hiding}
$\VC=(\VCGen,\VCCommit,\VCOpen,\VCCheck)$ has \defemph{hiding error $\VCHiding$} if for every security parameter $\Security\in\N$, message length $\VCLength\in\N$, public parameters $\VCParams\in\SupportOf{\VCGen(1^\Security,\VCLength)}$, query set $Q\subseteq[\VCLength]$, and messages $\VCVector^0,\VCVector^1\in\bits^{\VCLength}$ with $\VCVector^0[Q]=\VCVector^1[Q]$, the following two random variables are $\VCHiding(\Security,\VCLength)$-close in statistical distance:
\begin{itemize}[noitemsep]
  \item $\left\{
  (\VCCommitment,\VCOpening)
  \;\middle\vert\;
  \begin{array}{l}
    (\VCCommitment,\VCState)\gets\VCCommit(\VCParams,\VCVector^0)\\
    \VCOpening\gets\VCOpen(\VCParams,\VCState,Q)
  \end{array}
  \right\}$;
  \item $\left\{
  (\VCCommitment,\VCOpening)
  \;\middle\vert\;
  \begin{array}{l}
    (\VCCommitment,\VCState)\gets\VCCommit(\VCParams,\VCVector^1)\\
    \VCOpening\gets\VCOpen(\VCParams,\VCState,Q)
  \end{array}
  \right\}$.
\end{itemize}
\end{definition}

\begin{definition}
\label{def:succinct-vc}
$\VC=(\VCGen,\VCCommit,\VCOpen,\VCCheck)$ is \defemph{succinct} if the running time of $\VCGen$ and $\VCCheck$ and the lengths of $\VCParams$, $\VCCommitment$, and $\VCOpening$ are $\poly(\Security,\abs{Q},\log\VCLength)$, and $\VCCommit$ and $\VCOpen$ run in time $\poly(\Security,\VCLength)$.
\end{definition}

\begin{theorem}
\label{thm:vc-from-crh}
Let $\GeneralOracle$ be an oracle. If there exists a keyed hash family with collision-resistance error $\HashCollision$ relative to $\GeneralOracle$ (\cref{def:crh}), then there exists a succinct vector commitment scheme with perfect completeness, hiding error $\VCHiding(\Security,\VCLength)\leq2^{-\Security}$, and position binding error
\begin{equation*}
\VCBinding^{\GeneralOracleInstance}(\Security,\VCLength,s,\GenericQueryBound,\GenericTimeBound)
\leq
\HashCollision^{\GeneralOracleInstance}(\Security,\GenericQueryBound,\GenericTimeBound+\poly(\Security,s,\log\VCLength))
\end{equation*}
relative to $\GeneralOracle$; in particular, it is position binding relative to $\GeneralOracle$ if the family is collision resistant relative to $\GeneralOracle$.
\end{theorem}

\begin{proof}
Commit to each bit of the message with the statistically hiding commitment of Halevi--Micali~\cite{HM96}, with hiding parameter $\Security+\lceil\log\VCLength\rceil$, and place the $\VCLength$ commitments at the leaves of a Merkle tree~\cite{Merkle89} built from the hash family: two accepting openings that disagree at a position yield a collision in the hash family, and a hybrid over the leaves bounds the hiding error by $\VCLength\cdot2^{-\Security-\lceil\log\VCLength\rceil}\leq2^{-\Security}$.
\end{proof}

\doclearpage
\section{Impossibility of witness hiding in the random oracle model}
\label{sec:wh-lower-bound}
We prove that there are no witness-hiding relativized arguments for $\NTIME^\RandomOracle$:
\begin{theorem}
\label{thm:wh-lower-bound-rom}
Let $t\colon \N\to \N$ be a time-constructible function with $t(n)\geq n$ and $\RandomOracle$ be the random oracle over $((\bits^{t(n)+\lceil\log t(n)\rceil + 1},\bits^\Security))_{\Security,\InstanceSize\in\N}$. 
\begin{equation*}
\NTIME^{\RandomOracle}  \left[t\right]\not\subseteq\;
\IARG ^{\RandomOracle} 
\left[
\begin{array}{l l l l l l  }
  \emph{\textrm{completeness error}}	& 	\IARGCompleteness & = &1/10\\
  \emph{\textrm{soundness error}}	&	\IARGSoundness & = & \begin{cases}
    1/10 & 
      \IARGSoundnessAdvQueryBound=\poly(\Security,t(n))\enspace, \\&\IARGSoundnessAdvTimeBound=\poly(\Security,t(n))\\
    1 & \text{otherwise}\\
  \end{cases}  \\
  \emph{\textrm{verifier query bound}}	&	\IARGVerifierQuery & = & \poly(\Security,t(n))\\
  \emph{\textrm{verifier running time}}	&	\IARGVerifierTime & = & \poly(\Security,t(n))\\
  \emph{\textrm{honest prover query bound}}	&	\IARGProverQuery & = & \poly(\Security,t(n))\\
  \emph{\textrm{honest prover time bound}}	&	\IARGProverTime & = & \poly(\Security,t(n))\\
  \emph{\textrm{witness-hiding error}}	&	\IARGWitnessHiding & = & \begin{cases}
    1/10 & \WHAdversaryQuery=\poly(\Security,t(n))\enspace,
    \\&\WHAdversaryTime=\poly(\Security,t(n))\\
    1 & \text{otherwise}\\
  \end{cases}  \\
\end{array}
\right]\enspace.
\end{equation*}
\end{theorem}

In order to prove \cref{thm:wh-lower-bound-rom}, we prove that there are no witness-hiding relativized arguments for the relativized relation of \cite{BarbaraCG25}, which we recall next.
\begin{definition}[{\cite[Definition 5.3]{BarbaraCG25}}]
\label{def:relation-rom-ntime}
Let $t\colon \N\to \N$  be a time-constructible function with $t(n)\geq n$ for every $n\in \N$ and $\RandomOracle$ be the random oracle over $((\bits^{t(n)+\lceil\log t(n)\rceil + 1},\bits^\Security))_{\Security,\InstanceSize\in\N}$.
Let $\IndexEncoding{k}{i}$ denote the $(\lceil \log k\rceil+1)$-bit string that represents $i\in[k]$. We define $\Relation_{\RandomOracle}=\{\Relation_\ROInstance\}_{\Security,\InstanceSize\in\N, \ROInstance\in\RanOracleSec{\Security,\InstanceSize}}$ as follows.
\begin{equation*}
\Relation_\ROInstance\coloneq
\left\{
(\Instance,\Witness) \in \bits^n\times \bits^{t(n)}
\;\middle\vert\;
\begin{array}{l}
  \Instance=0^n\\
  \land\, \forall\, i\in[t(n)], \ROInstance(\Witness\concat \IndexEncoding{t(n)}{i})_1=0
\end{array}
\right\}\enspace.
\end{equation*}
\end{definition}

\begin{claim}[{\cite[Lemma 5.4]{BarbaraCG25}}]
\label{claim:the-rom-relation-ntime}
\begin{equation*}
  \Relation_{\RandomOracle}\in \NTIME^\RandomOracle[t]\enspace.
\end{equation*}
\end{claim}

\Cref{thm:wh-lower-bound-rom} follows from \cref{claim:the-rom-relation-ntime} and the following lemma since for every sufficiently large $\InstanceSize$,
\begin{equation*}
\left(1-\frac{1}{2^{t(\InstanceSize)}}\right)^{2^{t(\InstanceSize)}}
> 0.36
\enspace.
\end{equation*}

\begin{lemma}
\label{lem:wh-parameters-inequality}
Let $t\colon \N\to \N$  be a time-constructible function with $t(n)\geq n$ for every $n\in \N$ and $\RandomOracle$ be the random oracle over $((\bits^{t(n)+\lceil\log t(n)\rceil + 1},\bits^\Security))_{\Security,\InstanceSize\in\N}$. Let $\Relation_\RandomOracle$ be as in \cref{def:relation-rom-ntime}. If $\IARG = (\IARGProver,\IARGVerifier)$ is a relativized argument for $\Relation_{\RandomOracle}$, then the efficiency parameters of $\IARG$ satisfy
\begin{equation}
\label{eq:wh-parameters}
\IARGCompleteness
+
\IARGSoundness(\IARGProverQuery,O(\IARGProverTime+\IARGProverQuery\cdot t(\InstanceSize)))
+ 
\IARGWitnessHiding(O(\IARGVerifierQuery\cdot t(\InstanceSize)),O(\IARGVerifierTime+\IARGVerifierQuery\cdot t(\InstanceSize)))
\geq
\left(1-\frac{1}{2^{t(\InstanceSize)}}\right)^{2^{t(\InstanceSize)}}
\enspace,
\end{equation}
where we suppress the common security parameter $\Security$ and instance size $\InstanceSize$ for simplicity. 
\end{lemma}
We fix $\Security,\InstanceSize\in\N$ and suppress them throughout the rest of this section. The proof of \cref{lem:wh-parameters-inequality} relies on the fact that we can switch between completeness and soundness cases by modifying the oracle at a few locations. In \cref{sec:how-to-modify-the-oracle}, we define how we modify the oracle to insert a witness and show how this modification changes the probability distributions. In \cref{sec:proof-of-wh-lowerbound-lemma}, we prove \cref{lem:wh-parameters-inequality}.

\subsection{How to modify the oracle to insert a witness}
\label{sec:how-to-modify-the-oracle}
We define a function $\InsertOperation$ to modify the oracle to insert a witness as follows. For $\ROInstance\in\RandomOracle_{\Security,\InstanceSize}$, $\Witness\in\WitnessDomain$,
\begin{itemize}[nolistsep]
  \item[] $\Insert{\ROInstance}{\Witness}$:
  \begin{enumerate}[nolistsep]
    \item Initialize $\ROInstance'\coloneq \ROInstance$.
    \item For all $i\in[t(n)]$: $\ROInstance'(\Witness\concat \IndexEncoding{t(n)}{i})_1\coloneq 0$.
    \item Output $\ROInstance'$.
  \end{enumerate} 
\end{itemize}

We note that $\InsertOperation$ is identical to how \cite{BarbaraCG25} modifies the oracle and the following claims are implicit in \cite{BarbaraCG25}.
\begin{lemma}
\label{lem:how-modifying-oracle-changes-distributions}
For every set $\GenericOracleAlgorithmInputSet$, predicate $\GenericPredicate$ over $\RandomOracle_{\Security,\InstanceSize}\times(\WitnessDomain\cup\{\bot\})\times\GenericOracleAlgorithmInputSet$, and family of distributions $\{\Distribution_{({\ROInstanceWithFirstWitness},{\Witness_0})}\}_{({\ROInstanceWithFirstWitness},{\Witness_0})\in\RandomOracle_{\Security,\InstanceSize}\times(\WitnessDomain\cup\{\bot\})}$ over $\GenericOracleAlgorithmInputSet$,
\begin{equation*}
\probOverExp{\RandomExperiment}{
\GenericPredicate(\ROInstanceWithFirstWitness, \Witness_0, \GenericOracleAlgorithmInput)=1
\,\land\, 0^\InstanceSize \notin \GetLanguage{\Relation_\ROInstance}  
}
=
\left(
  1-\frac{1}{2^{t(\InstanceSize)}}
\right)
\cdot
\probOverExp{\RandomExperiment'}{
\GenericPredicate(\ROInstance, \Witness_0, \GenericOracleAlgorithmInput)=1
\,\land\,(0^n,\Witness_0) \in \Relation_\ROInstance
}
\enspace,
\end{equation*}
where $\IARGCompletenessAdv$ is defined as follows.
\begin{itemize}[nolistsep]
  \item[] $\IARGCompletenessAdv^\ROInstance\colon$
  \begin{enumerate}[nolistsep]
  \item Compute $\WitnessesOf{\ROInstance}{0^\InstanceSize}{\Relation_\RandomOracle}$ by querying $\ROInstance$ as needed.
  \item If there exists some $\Witness$ such that $\WitnessesOf{\ROInstance}{0^\InstanceSize}{\Relation_\RandomOracle} = \{\Witness\}$, output $\Witness$. Otherwise, output $\bot$.
  \end{enumerate} 
\end{itemize}
and
\begin{equation*}
\RandomExperiment
\coloneq
\left\{ 
\begin{array}{l}
  \ROInstance\gets \RandomOracle_{\Security,\InstanceSize}\\
  \Witness_0\gets \WitnessDomain\\
  \ROInstanceWithFirstWitness \coloneq \Insert{\ROInstance}{\Witness_0}\\
  \GenericOracleAlgorithmInput \gets \Distribution_{({\ROInstanceWithFirstWitness},{\Witness_0})}
\end{array}
\right\}
\quad\text{and}\quad
\RandomExperiment'
\coloneq
\left\{ 
\begin{array}{l}
  \ROInstance\gets \RandomOracle_{\Security,\InstanceSize}\\
  \Witness_0\gets \IARGCompletenessAdv^\ROInstance\\
  \GenericOracleAlgorithmInput \gets \Distribution_{({\ROInstance},{\Witness_0})}
\end{array}
\right\}
\enspace.
\end{equation*}
\end{lemma}
\begin{proof}
We have
\begin{align*}
\probOverExp{\RandomExperiment}{
\GenericPredicate(\ROInstanceWithFirstWitness, \Witness_0, \GenericOracleAlgorithmInput)=1
\,\land\, 0^\InstanceSize \notin \GetLanguage{\Relation_\ROInstance}  
}
&=
\probOverExp{\RandomExperiment}{
\GenericPredicate(\ROInstanceWithFirstWitness, \Witness_0, \GenericOracleAlgorithmInput)=1
\,\land\, 0^\InstanceSize \notin \GetLanguage{\Relation_\ROInstance}  
}
\\&\quad-
\probOverExp{\RandomExperiment}{
\begin{array}{l}
\GenericPredicate(\ROInstanceWithFirstWitness, \Witness_0, \GenericOracleAlgorithmInput)=1
\,\land\, 0^\InstanceSize \notin \GetLanguage{\Relation_\ROInstance}  \\
\land\, \WitnessesOf{\ROInstanceWithFirstWitness}{0^\InstanceSize}{\Relation_\RandomOracle} \neq \{\Witness_0\}
\end{array}
}
\tag{since $\InsertOperation$ adds exactly $\Witness$ to the witness set by definition}
\\
&=
\probOverExp{\RandomExperiment}{
\begin{array}{l}
\GenericPredicate(\ROInstanceWithFirstWitness, \Witness_0, \GenericOracleAlgorithmInput)=1
\,\land\, 0^\InstanceSize \notin \GetLanguage{\Relation_\ROInstance}\\
\land\, \WitnessesOf{\ROInstanceWithFirstWitness}{0^\InstanceSize}{\Relation_\RandomOracle} = \{\Witness_0\}
\end{array}
}
\tag{by the law of total probability}
\enspace.
\end{align*}
Therefore, we are left to prove
\begin{equation}
\label{eq:distribution-of-modified-oracle-main-eq}
\probOverExp{\RandomExperiment}{
\begin{array}{l}
\GenericPredicate(\ROInstanceWithFirstWitness, \Witness_0, \GenericOracleAlgorithmInput)=1
\,\land\, 0^\InstanceSize \notin \GetLanguage{\Relation_\ROInstance}\\
\land\, \WitnessesOf{\ROInstanceWithFirstWitness}{0^\InstanceSize}{\Relation_\RandomOracle} = \{\Witness_0\}
\end{array}
}
=
\left(
  1-\frac{1}{2^{t(\InstanceSize)}}
\right)
\cdot
\probOverExp{\RandomExperiment'}{
\begin{array}{l}
\GenericPredicate(\ROInstance, \Witness_0, \GenericOracleAlgorithmInput)=1\\
\land\,(0^n,\Witness_0) \in \Relation_\ROInstance
\end{array}
}
\enspace.
\end{equation}

By the law of total probability and chain rule, the left-hand side of \Cref{eq:distribution-of-modified-oracle-main-eq} equals
\begin{equation*}
\sum_{
  \substack{
  \Sampled{\ROInstanceWithFirstWitness} \in \RandomOracle_{\Security,\InstanceSize}, \Sampled{\Witness_0} \in \WitnessDomain\\
  \text{s.t. } \WitnessesOf{\Sampled{\ROInstanceWithFirstWitness}}{0^\InstanceSize}{\Relation_\RandomOracle} = \{\Sampled{\Witness_0}\}
  }
}
\probOverExp{\RandomExperiment}{
  \begin{array}{l}
  \ROInstanceWithFirstWitness = \Sampled{\ROInstanceWithFirstWitness}\\
  \land\,
  \Witness_0 = \Sampled{\Witness_0} \\
  \land\, 0^\InstanceSize \notin \GetLanguage{\Relation_\ROInstance}\\   
  \end{array}
}\cdot
\prob{
\GenericPredicate(\Sampled{\ROInstanceWithFirstWitness}, \Sampled{\Witness_0}, \GenericOracleAlgorithmInput)=1
\;\middle\vert\;
  \GenericOracleAlgorithmInput \gets \Distribution_{(\Sampled{\ROInstanceWithFirstWitness},\Sampled{\Witness_0})}
}
\enspace,
\end{equation*} 
while the right-hand side of \Cref{eq:distribution-of-modified-oracle-main-eq} equals
\begin{equation*}
\sum_{
  \substack{
  \Sampled{\ROInstanceWithFirstWitness} \in \RandomOracle_{\Security,\InstanceSize}, \Sampled{\Witness_0} \in \WitnessDomain\\
  \text{s.t. } \WitnessesOf{\Sampled{\ROInstanceWithFirstWitness}}{0^\InstanceSize}{\Relation_\RandomOracle} = \{\Sampled{\Witness_0}\}
  }
}
\left(
  1-\frac{1}{2^{t(\InstanceSize)}}
\right)
\cdot
\probOverExp{\RandomExperiment'}{
\begin{array}{l}
\ROInstance = \Sampled{\ROInstanceWithFirstWitness}\\
\land\, \Witness_0 = \Sampled{\Witness_0}\\
\land\,(0^n,\Witness_0) \in \Relation_\ROInstance
\end{array}
}
\cdot
\prob{
\GenericPredicate(\Sampled{\ROInstanceWithFirstWitness}, \Sampled{\Witness_0}, \GenericOracleAlgorithmInput)=1
\;\middle\vert\;
  \GenericOracleAlgorithmInput \gets \Distribution_{(\Sampled{\ROInstanceWithFirstWitness},\Sampled{\Witness_0})}
}
\enspace.
\end{equation*}
Consequently, proving that for every $\Sampled{\Witness_0}\in \WitnessDomain$, $\Sampled{\ROInstanceWithFirstWitness}\in \RandomOracle_{\Security,\InstanceSize}$ such that $\WitnessesOf{\Sampled{\ROInstanceWithFirstWitness}}{0^\InstanceSize}{\Relation_\RandomOracle} = \{\Sampled{\Witness_0}\}$,
\begin{equation}
\label{eq:changing-distributions}
\probOverExp{\RandomExperiment}{
\begin{array}{l}
\ROInstanceWithFirstWitness = \Sampled{\ROInstanceWithFirstWitness}
\,\land\, \Witness_0 = \Sampled{\Witness_0}\\
\land\, 0^\InstanceSize \notin \GetLanguage{\Relation_\ROInstance}\\
\end{array}
}
=
\left(
1-\frac{1}{2^{t(\InstanceSize)}}
\right)
\cdot
\probOverExp{\RandomExperiment'}{
\begin{array}{l}
\ROInstance = \Sampled{\ROInstanceWithFirstWitness}
\,\land\, \Witness_0 = \Sampled{\Witness_0}\\
\land\,(0^n,\Witness_0) \in \Relation_\ROInstance
\end{array}
}
\enspace,
\end{equation}
proves \Cref{eq:distribution-of-modified-oracle-main-eq}. In the rest, we fix $\Sampled{\Witness_0}\in \WitnessDomain$, $\Sampled{\ROInstanceWithFirstWitness}\in \RandomOracle_{\Security,\InstanceSize}$ such that $\WitnessesOf{\Sampled{\ROInstanceWithFirstWitness}}{0^\InstanceSize}{\Relation_\RandomOracle} = \{\Sampled{\Witness_0}\}$ and prove \Cref{eq:changing-distributions}. Let 
\begin{equation*}
\SummationDomain 
\coloneq
\{
  \Sampled{\ROInstance} \in \RandomOracle_{\Security,\InstanceSize}
  \colon
  \Sampled{\ROInstanceWithFirstWitness}=\Insert{\Sampled{\ROInstance}}{\Sampled{\Witness_0}}
  \,\land\,
  0^\InstanceSize \notin \GetLanguage{\Relation_{\Sampled{\ROInstance}}}
  \}
\enspace.
\end{equation*} 
We have
\begin{align*}
\probOverExp{\RandomExperiment}{
\begin{array}{l}
\ROInstanceWithFirstWitness = \Sampled{\ROInstanceWithFirstWitness}
\,\land\, \Witness_0 = \Sampled{\Witness_0}\\
\land\, 0^\InstanceSize\not\in \GetLanguage{\Relation_\ROInstance}
\end{array}
}
&=
\sum_{
  \substack{
    \Sampled{\ROInstance} \in \SummationDomain
  }
}
\probOverExp{\RandomExperiment}{
\begin{array}{l}
\ROInstance = \Sampled{\ROInstance}
\,\land\, \Witness_0 = \Sampled{\Witness_0}\\
\land\,
0^\InstanceSize\not\in \GetLanguage{\Relation_\ROInstance}
\end{array}
}
\tag{by the law of total probability}
\\
&=
\sum_{
  \substack{
    \Sampled{\ROInstance} \in \SummationDomain
  }
}
\probOverExp{\RandomExperiment}{
\ROInstance = \Sampled{\ROInstance}
\,\land\, \Witness_0 = \Sampled{\Witness_0}
}
\tag{by the definition of $\SummationDomain$}
\\
&=
\sum_{
  \substack{
    \Sampled{\ROInstance} \in \SummationDomain
  }
}
\probOverExp{\RandomExperiment}{
\ROInstance = \Sampled{\ROInstanceWithFirstWitness}
\,\land\, \Witness_0 = \Sampled{\Witness_0}
}
\tag{by uniformity of $\RandomOracle$}
\\
&=
\frac{1}{2^{t(\InstanceSize)}}
\cdot
\sum_{
  \substack{
    \Sampled{\ROInstance} \in \SummationDomain
  }
}
\prob{
\ROInstance = \Sampled{\ROInstanceWithFirstWitness}
\;\middle\vert\;
\ROInstance\gets \RandomOracle_{\Security,\InstanceSize}
}
\tag{by how $\ROInstance$ and $\Witness_0$ are sampled in $\RandomExperiment$}
\\
&=
\frac{\abs{\SummationDomain}}{2^{t(\InstanceSize)}}
\cdot
\prob{
\ROInstance = \Sampled{\ROInstanceWithFirstWitness}
\;\middle\vert\;
\ROInstance\gets \RandomOracle_{\Security,\InstanceSize}
}
\tag{by summing over $\Sampled{\ROInstance}$}
\\
&=
\frac{\abs{\SummationDomain}}{2^{t(\InstanceSize)}}
\cdot
\frac{
\probOverExp{\RandomExperiment'}{
\ROInstance = \Sampled{\ROInstanceWithFirstWitness}
\,\land\,
\Witness_0 = \Sampled{\Witness_0}
}
}
{
\prob{
\Witness_0 = \Sampled{\Witness_0}
\;\middle\vert\;
\Witness_0 \gets \IARGCompletenessAdv^{\Sampled{\ROInstanceWithFirstWitness}}
}
}
\tag{by the chain rule}
\\
&=
\frac{\abs{\SummationDomain}}{2^{t(\InstanceSize)}}
\cdot
\probOverExp{\RandomExperiment'}{
\begin{array}{l}
\ROInstance = \Sampled{\ROInstanceWithFirstWitness}
\,\land\,
\Witness_0 = \Sampled{\Witness_0}\\
\land\, (0^n,\Witness_0) \in \Relation_\ROInstance
\end{array}
}
\tag{since $\WitnessesOf{\Sampled{\ROInstanceWithFirstWitness}}{0^\InstanceSize}{\Relation_\RandomOracle}=\{\Sampled{\Witness_0}\}$}
\enspace.
\end{align*}
Finally, we have $\abs{\SummationDomain} = 2^{t(\InstanceSize)}-1$ because
\begin{equation*}
\abs{\SummationDomain}
=
\abs{\{
  \Recover{\Sampled{\ROInstanceWithFirstWitness}}{\Sampled{\Witness_0}}{\BitsForWitness}
  \colon
  \BitsForWitness\in\bits^{t(n)}\setminus \{0^{t(\InstanceSize)}\}
\}}
=
2^{t(\InstanceSize)}-1
\enspace,
\end{equation*}
where $\RecoverOperation$ is defined as follows for $\ROInstance\in\RandomOracle_{\Security,\InstanceSize}$, $\Witness\in\bits^{t(\InstanceSize)}$, $\BitsForWitness\in\bits^{t(n)}$, 
\begin{itemize}[nolistsep]
  \item[] $\Recover{\ROInstance}{\Witness}{\BitsForWitness}$:
  \begin{enumerate}[nolistsep]
    \item Initialize $\ROInstance'\coloneq \ROInstance$.
    \item For all $i\in[t(n)]$: $\ROInstance'(\Witness\concat \IndexEncoding{t(n)}{i})_1\coloneq \BitsForWitness_i$.
    \item Output $\ROInstance'$.
  \end{enumerate} 
\end{itemize}

\end{proof}

\subsection{Proof of \cref{lem:wh-parameters-inequality}}
\label{sec:proof-of-wh-lowerbound-lemma}
Throughout this proof, we sample the random variables according to the following random experiment.
\begin{equation*}
\RandomExperiment\coloneq
\left\{
\begin{array}{l} 
  \ROInstance\gets \RandomOracle_{\Security,\InstanceSize}\\
  \Witness_0 \gets \WitnessDomain\\
  \ROInstanceWithFirstWitness \coloneq \Insert{\ROInstance}{\Witness_0}\\
  \IARGDecision \xleftarrow{\IARGProverRandomness,\IARGVerifierRandomness,(\IARGProverMessage_{1},\IARGVerifierMessage_{1},\dots,\IARGProverMessage_{\IARGRoundComplexity},\IARGVerifierMessage_{\IARGRoundComplexity})} \langle \IARGProver^{\ROInstanceWithFirstWitness}(1^\Security,(0^\InstanceSize,\Witness_0)),\IARGVerifier^{\ROInstance}(1^\Security,0^\InstanceSize)\rangle \\
  \IARGDecisionForFirstWitness \xleftarrow{(\IARGPMsgForFirstWitness{1},\IARGVMsgForFirstWitness{1},\dots,\IARGPMsgForFirstWitness{\IARGRoundComplexity},\IARGVMsgForFirstWitness{\IARGRoundComplexity})} \langle \IARGProver^{\ROInstanceWithFirstWitness}(1^\Security,(0^n,\Witness_0);\IARGProverRandomness),\IARGVerifier^{\ROInstanceWithFirstWitness}(1^\Security,0^n;\IARGVerifierRandomness)\rangle \\
\end{array}
\right\}\enspace.
\end{equation*}

To prove \Cref{eq:wh-parameters}, we analyze the probability that the verifier queries the oracle at a location relevant for $\Witness_0$. Formally, we define a verifier query predicate as follows: $\VQueryPredicate{\ROInstance}{(\IARGProverMessage_i)_{i\in[\IARGRoundComplexity]}}{\IARGVerifierRandomness}{\Witness}$ evaluates to $1$ if and only if the verifier $\IARGVerifier^\ROInstance(1^\Security,0^\InstanceSize,(\IARGProverMessage_i)_{i\in[j-1]};\IARGVerifierRandomness)$ queries 
$\ROInstance$ at $\Witness\concat\IndexEncoding{t(\InstanceSize)}{i}$ for some $i\in\WitnessIndexRange{n}$ and $j\in[\IARGRoundComplexity+1]$.

\Cref{eq:wh-parameters} follows from the following.
\begin{align*}
\left(1-\frac{1}{2^{t(\InstanceSize)}}\right)^{2^{t(\InstanceSize)}}
- \IARGSoundness(\IARGProverQuery,O(\IARGProverTime+t(\InstanceSize)\cdot\IARGProverQuery))
- \IARGCompleteness
&=
\prob{0^\InstanceSize \notin \GetLanguage{\Relation_\ROInstance}} 
- \IARGSoundness(\IARGProverQuery,O(\IARGProverTime+t(\InstanceSize)\cdot\IARGProverQuery))
- \IARGCompleteness
\tag{by \cref{claim:the-rom-relation-weight}, proven below}
\\
&\leq
\prob{
  \VQueryPredicate{\ROInstanceWithFirstWitness}{(\IARGPMsgForFirstWitness{i})_{i\in[\IARGRoundComplexity]}}{\IARGVerifierRandomness}{\Witness_0}=1
  \,\land\,0^\InstanceSize \notin \GetLanguage{\Relation_\ROInstance}
}
\tag{by \cref{claim:total-of-completeness-soundness-and-QV-is-constant}, proven below}
\\
&\leq
\IARGWitnessHiding(O(\IARGVerifierQuery\cdot t(\InstanceSize)),O(\IARGVerifierTime+\IARGVerifierQuery\cdot t(\InstanceSize)+ \IARGVerifierQuery \cdot t(\InstanceSize)))
\tag{by \cref{claim:wh-main-lower-bound}, proven below}
\enspace.
\end{align*}

We finish our proof by proving the missing claims.

\begin{claim}
\label{claim:the-rom-relation-weight}
\begin{align*}
\prob{
  0^\InstanceSize \notin \GetLanguage{\Relation_{\ROInstance}}
}
&=
\left(1-\frac{1}{2^{t(\InstanceSize)}}\right)^{2^{t(\InstanceSize)}}
\enspace.
\end{align*}
\end{claim}
\begin{proof}
  We have
\begin{align*}
\prob{
  0^\InstanceSize \notin \GetLanguage{\Relation_{\ROInstance}}
}
&=
\prob{
  \forall\, \Witness \in \WitnessDomain,  
  (0^\InstanceSize,\Witness)\notin\Relation_{\ROInstance}
}
\tag{by \cref{def:relativized-relation}}
\\
&=
\prod_{\Witness \in \WitnessDomain}
\prob{ 
  (0^\InstanceSize,\Witness)\notin\Relation_{\ROInstance}
}
\tag{by \cref{def:relation-rom-ntime}, the events $\{(0^\InstanceSize,\Witness)\in\Relation_{\ROInstance}\}_{\Witness\in\WitnessDomain}$ are mutually independent}
\\
&=
\prod_{\Witness \in \WitnessDomain}
\left(1-\frac{1}{2^{t(\InstanceSize)}}\right)
\tag{$(0^n,\Witness)\in\Relation_{\ROInstance}$ iff for every $i\in[t(\InstanceSize)]$, $\ROInstance(\Witness\concat\IndexEncoding{t(\InstanceSize)}{i})_1=0$}
=
\left(1-\frac{1}{2^{t(\InstanceSize)}}\right)^{2^{t(\InstanceSize)}}
\enspace.
\end{align*}
\end{proof}

\begin{claim}
\label{claim:total-of-completeness-soundness-and-QV-is-constant}
\begin{equation*}
\prob{
  \VQueryPredicate{\ROInstanceWithFirstWitness}{(\IARGPMsgForFirstWitness{i})_{i\in[\IARGRoundComplexity]}}{\IARGVerifierRandomness}{\Witness_0}=1
  \,\land\,0^\InstanceSize \notin \GetLanguage{\Relation_\ROInstance}
}
\geq
\prob{0^\InstanceSize \notin \GetLanguage{\Relation_\ROInstance}} - \IARGSoundness(\IARGProverQuery,O(\IARGProverTime+t(\InstanceSize)\cdot\IARGProverQuery))
-
\IARGCompleteness
\enspace.
\end{equation*}
\end{claim}
\begin{proof}
We define a malicious prover $\IARGSoundnessAdv$ as follows.
\begin{itemize}
  \item[] $\IARGSoundnessAdv^\ROInstance(\Instance)$:
  \begin{enumerate}[nolistsep]
    \item Sample $\IARGProverRandomness\gets\bits^\IARGProverRandomnessComplexity$.
    \item Sample $\Witness_0 \gets \WitnessDomain$.
    \item For $i\in[\IARGRoundComplexity]$:
    \begin{enumerate}[nolistsep]
    \item Simulate $\IARGProverMessage_i \coloneq \IARGProver^{\Insert{\ROInstance}{\Witness_0}}(1^\Security,(\Instance,\Witness_0),(\IARGVerifierMessage_j)_{j\in[i-1]};\IARGProverRandomness)$ by answering each oracle query $\OracleQuery$ of $\IARGProver$ as follows.
    \begin{enumerate}[nolistsep]
    \item Compute $\OracleAnswer \coloneq \Insert{\ROInstance}{\Witness_0}(\OracleQuery)$ as follows.
    \begin{enumerate}
    \item Query the oracle at $\OracleQuery$: $\OracleAnswer\coloneq\ROInstance(\OracleQuery)$.
    \item Parse $\OracleQuery$ as $(\Witness',\IndexEncoding{t(n)}{i})$.
    \item If $\Witness'=\Witness_0$ and $i\in[t(n)]$, set $\OracleAnswer_1\coloneq 0$. 
    \end{enumerate}
    \item Answer $\IARGProver$'s oracle call with $\OracleAnswer$.
    \end{enumerate}
    \item Output $\IARGProverMessage_i$.
    \item Receive the $i$-th verifier message $\IARGVerifierMessage_i$.
    \end{enumerate}
  \end{enumerate} 
\end{itemize}
Answering an oracle query of $\IARGSoundnessAdv$ takes time $O(t(\InstanceSize))$ and $1$ query to the oracle. Therefore, $\IARGSoundnessAdv$ has the running time $O(\IARGProverTime+\IARGProverQuery\cdot t(\InstanceSize))$ and makes at most $\IARGProverQuery$ queries to the oracle. Note that $\IARGProver^{\ROInstanceWithFirstWitness}(1^\Security,(0^\InstanceSize,\Witness_0))$ in $\RandomExperiment$ is equivalent to $\IARGSoundnessAdv^\ROInstance(0^\InstanceSize)$.

We have 
\begin{align*}
&\prob{
  0^\InstanceSize \notin \GetLanguage{\Relation_\ROInstance}
}
- \IARGSoundness(\Security,n,\IARGProverQuery,O(\IARGProverTime+\IARGProverQuery\cdot t(\InstanceSize)))
\\
&\leq 
\prob{
  0^\InstanceSize \notin \GetLanguage{\Relation_\ROInstance}
}
-
\prob{
  \IARGDecision = 1
  \,\land\,0^\InstanceSize \notin \GetLanguage{\Relation_\ROInstance}    
}
\tag{by the soundness of $\IARG$}
\\
&=
\prob{
  \IARGDecision = 0
  \,\land\,0^\InstanceSize \notin \GetLanguage{\Relation_\ROInstance}    
}
\tag{by the law of total probability}
\\
&=
\prob{
\begin{array}{l}
\IARGDecision = 0
\,\land\, 
0^\InstanceSize \notin \GetLanguage{\Relation_\ROInstance}
\\\land\,
\VQueryPredicate{\ROInstanceWithFirstWitness}{(\IARGPMsgForFirstWitness{i})_{i\in[\IARGRoundComplexity]}}{\IARGVerifierRandomness}{\Witness_0}=1
\end{array}
}
+
\prob{
\begin{array}{l}
\IARGDecision = 0
\,\land\, 
0^\InstanceSize \notin \GetLanguage{\Relation_\ROInstance}
\\\land\,
\VQueryPredicate{\ROInstanceWithFirstWitness}{(\IARGPMsgForFirstWitness{i})_{i\in[\IARGRoundComplexity]}}{\IARGVerifierRandomness}{\Witness_0}=0
\end{array}  
}
\tag{by the law of total probability}
\\
&=
\prob{
\begin{array}{l}
\IARGDecision = 0
\,\land\, 
0^\InstanceSize \notin \GetLanguage{\Relation_\ROInstance}
\\\land\,
\VQueryPredicate{\ROInstanceWithFirstWitness}{(\IARGProverMessage'_i)_{i\in[\IARGRoundComplexity]}}{\IARGVerifierRandomness}{\Witness_0}=1
\end{array}  
}
+
\prob{
\begin{array}{l}
\IARGDecisionForFirstWitness = 0
\,\land\, 
0^\InstanceSize \notin \GetLanguage{\Relation_\ROInstance}
\\\land\,
\VQueryPredicate{\ROInstanceWithFirstWitness}{(\IARGPMsgForFirstWitness{i})_{i\in[\IARGRoundComplexity]}}{\IARGVerifierRandomness}{\Witness_0}=0
\end{array}
}
\tag{if $\VQueryPredicate{\ROInstanceWithFirstWitness}{(\IARGPMsgForFirstWitness{i})_{i\in[\IARGRoundComplexity]}}{\IARGVerifierRandomness}{\Witness_0}=0$, then $\IARGDecision=\IARGDecisionForFirstWitness$ because the verifier's view is the same}
\\
&\leq
\prob{
0^\InstanceSize \notin \GetLanguage{\Relation_\ROInstance}
\,\land\,
\VQueryPredicate{\ROInstanceWithFirstWitness}{(\IARGPMsgForFirstWitness{i})_{i\in[\IARGRoundComplexity]}}{\IARGVerifierRandomness}{\Witness_0}=1
}
+
\prob{
\IARGDecisionForFirstWitness = 0
\,\land\, 
0^\InstanceSize \notin \GetLanguage{\Relation_\ROInstance}
}
\tag{since omitting an event can only increase the probability}
\enspace.
\end{align*}
We are left to prove that
\begin{equation}
\label{eq:bound-with-completeness}
\prob{
\IARGDecisionForFirstWitness = 0
\,\land\, 
0^\InstanceSize \notin \GetLanguage{\Relation_\ROInstance}
}
\leq
\IARGCompleteness
\enspace.  
\end{equation}

Let the completeness adversary $\IARGCompletenessAdv(\Instance)$ be as in \Cref{lem:how-modifying-oracle-changes-distributions}. The left-hand side of \cref{eq:bound-with-completeness} is at most
\begin{equation}
\label{eq:lemma-bound-with-completeness-proof}
\prob{
\IARGDecision = 0
\,\land\,
(0^\InstanceSize,\Witness_0) \in \Relation_\ROInstance
}
\end{equation} 
by \Cref{lem:how-modifying-oracle-changes-distributions} with $\GenericOracleAlgorithmInputSet = \bits^\IARGVerifierRandomnessComplexity\times\bits^{\IARGProverComm}$, $\GenericPredicate(\ROInstance,\Witness_0,\GenericOracleAlgorithmInput=(\IARGVerifierRandomness,(\IARGProverMessage_i)_{i\in[\IARGRoundComplexity]})) = 1-\IARGVerifier^{\ROInstance}(1^\Security,0^\InstanceSize,(\IARGProverMessage_i)_{i\in[\IARGRoundComplexity]};\IARGVerifierRandomness)$, 
\begin{equation*}
\Distribution_{(\ROInstance,\Witness_0)} 
=
\left\{
  \GenericOracleAlgorithmInput=(\IARGVerifierRandomness,(\IARGProverMessage_i)_{i\in[\IARGRoundComplexity]})
  \;\middle\vert\;
  \IARGDecision \xleftarrow{\IARGProverRandomness_0,\IARGVerifierRandomness,(\IARGProverMessage_1,\IARGVerifierMessage_1,\dots,\IARGVerifierMessage_\IARGRoundComplexity)} \langle \IARGProver^{\Insert{\ROInstance}{\Witness_0}}(1^\Security,(0^n,\Witness_0)),\IARGVerifier^{\ROInstance}(1^\Security,0^\InstanceSize) \rangle 
\right\}
\end{equation*}
for $(\ROInstance,\Witness_0)\in\RandomOracle_{\Security,\InstanceSize}\times(\WitnessDomain\cup\{\bot\})$.
We conclude the proof by observing that \Cref{eq:lemma-bound-with-completeness-proof} is at most $\IARGCompleteness$ by the completeness of $\IARG$.
\end{proof}

\begin{claim}
\label{claim:wh-main-lower-bound}
\begin{equation}
\label{eq:wh-main-lowerbound}
\prob{
\VQueryPredicate{\ROInstanceWithFirstWitness}{(\IARGPMsgForFirstWitness{i})_{i\in[\IARGRoundComplexity]}}{\IARGVerifierRandomness}{\Witness_0}=1
\,\land\,
0^\InstanceSize \notin \GetLanguage{\Relation_\ROInstance}
}
\leq
\IARGWitnessHiding(O(\IARGVerifierQuery\cdot t(\InstanceSize)),O(\IARGVerifierTime+\IARGVerifierQuery\cdot t(\InstanceSize)))
\enspace.
\end{equation} 
\end{claim}
\begin{proof}
We use the witness-hiding property of $\IARG$ to prove the claim. We define witness-hiding adversaries $\WHSampleAdversary$ and $\WHOutAdversary$ as follows. Let $M$ be the ND-decider for $\Relation_\RandomOracle$ that runs in time $O(t(\InstanceSize))$.
\begin{itemize}[noitemsep]
\item $\WHSampleAdversary^\ROInstance(\Instance)$:
  \begin{enumerate}[noitemsep]
  \item Compute $\WitnessesOf{\ROInstance}{0^\InstanceSize}{\Relation_\RandomOracle}$ by querying $\ROInstance$ as needed.
  \item If $\WitnessesOf{\ROInstance}{0^\InstanceSize}{\Relation_\RandomOracle} = \{\Witness_0\}$ for some $\Witness_0$, output $\Witness_0$.
  \item Otherwise, output $\bot$.
  \end{enumerate} 
\item $\WHOutAdversary^\ROInstance(\Instance,\IARGVerifierRandomness,\IARGTranscript)$:
  \begin{enumerate}[noitemsep]
  \item Parse $\IARGTranscript$ as $(\IARGProverMessage_1,\IARGVerifierMessage_1,\dots,\IARGProverMessage_\IARGRoundComplexity,\IARGVerifierMessage_\IARGRoundComplexity)$
  \item Set $S\coloneq \emptyset$.
  \item For $i\in\{1,\dots,\IARGRoundComplexity\}$, simulate $\IARGVerifierMessage_{i} \coloneq \IARGVerifier^{\ROInstance}(1^\Security,0^\InstanceSize,(\IARGProverMessage_{j})_{j\in[i]};\IARGVerifierRandomness)$ by answering each oracle query $\OracleQuery$ of $\IARGVerifier$ as follows.
    \begin{enumerate}[nolistsep]
    \item Query $\ROInstance$ at $\OracleQuery$; let $\OracleAnswer$ be the answer.
    \item Answer $\IARGVerifier$'s oracle call with $\OracleAnswer$.
    \item Parse $\OracleQuery$ as $(\Witness,\IndexEncoding{t(n)}{i})$ and set $S \coloneq S\cup\{\Witness\}$. 
    \end{enumerate}
  \item For $\Witness \in S$:
    \begin{enumerate}[nolistsep]
    \item Simulate $\out\coloneq M^\ROInstance(\Instance,\Witness)$ by querying the oracle at $\OracleQuery$ and answering $M$'s oracle query with  $\ROInstance(\OracleQuery)$ whenever $M$ queries the oracle at some $\OracleQuery$. 
    \item If $\out = 1$, output $\Witness$.
    \end{enumerate}
  \item Output $\bot$.
  \end{enumerate}
\end{itemize}
We now compute the time and query complexity of $\WHOutAdversary$. Answering an oracle query of $\IARGVerifier$ requires $1$ query to the oracle and $O(t(\InstanceSize))$ time. Running the ND-decider once requires $O(t(\InstanceSize))$ queries to the oracle and $O(t(\InstanceSize))$ time. Since the verifier makes at most $\IARGVerifierQuery$ queries to the oracle, $\WHOutAdversary$ makes at most $O(\IARGVerifierQuery\cdot t(\InstanceSize))$ oracle queries and runs in time $O(\IARGVerifierTime+\IARGVerifierQuery\cdot  t(\InstanceSize))$. Therefore,
\begin{equation}
\label{eq:wh-prob}
\begin{aligned}
&\prob{
\begin{array}{l}
  (0^\InstanceSize,\Witness') \in \Relation_\ROInstance
  \,\land\, (0^\InstanceSize,\Witness_0)\in\Relation_\ROInstance
\end{array}
\;\middle\vert\;
\begin{array}{l}
  \ROInstance\gets \RandomOracle_{\Security,\InstanceSize} \\
  \Witness_0\gets \WHSampleAdversary^{\ROInstance}(0^\InstanceSize) \\
  \IARGDecision \xleftarrow{\IARGProverRandomness,\IARGVerifierRandomness,\IARGTranscript} \langle \IARGProver^{\ROInstance}(1^\Security,(0^\InstanceSize,\Witness_0)), \IARGVerifier^{\ROInstance}(1^\Security,0^\InstanceSize) \rangle\\
  \Witness' \gets \WHOutAdversary^\ROInstance(0^\InstanceSize,\IARGVerifierRandomness,\IARGTranscript)
\end{array}
}
\end{aligned}
\end{equation}
is at most $\IARGWitnessHiding(\Security,\InstanceSize,O(\IARGVerifierQuery\cdot t(\InstanceSize)),O(\IARGVerifierTime+\IARGVerifierQuery \cdot t(\InstanceSize)))$. We are left to show that \Cref{eq:wh-prob} upper bounds
\begin{equation}
\label{eq:wh-proof-query-exp-for-modified-oracle} 
\probOverExp{\RandomExperiment}{
\VQueryPredicate{\ROInstanceWithFirstWitness}{(\IARGPMsgForFirstWitness{i})_{i\in[\IARGRoundComplexity]}}{\IARGVerifierRandomness}{\Witness_0}=1
\,\land\, 0^\InstanceSize\notin\GetLanguage{\Relation_\ROInstance}
}  
\enspace.
\end{equation}
\Cref{eq:wh-proof-query-exp-for-modified-oracle} equals
\begin{equation}
\label{eq:wh-proof-query-exp-for-original-oracle} 
\left(
  \frac{2^{t(\InstanceSize)}-1}{2^{t(\InstanceSize)}}
\right)
\cdot
\prob{
\begin{array}{l}
  \VQueryPredicate{\ROInstance}{(\IARGProverMessage_{i})_{i\in[\IARGRoundComplexity]}}{\IARGVerifierRandomness}{\Witness_0}=1\\
  \land\, (0^\InstanceSize,\Witness_0)\in\Relation_\ROInstance
\end{array}
\;\middle\vert\;
\begin{array}{l}
  \ROInstance\gets \RandomOracle_{\Security,\InstanceSize} \\
  \Witness_0\gets \WHSampleAdversary^{\ROInstance}(0^\InstanceSize) \\
  \IARGDecision \xleftarrow{\IARGProverRandomness,\IARGVerifierRandomness,\IARGTranscript} \langle \IARGProver^{\ROInstance}(1^\Security,(0^\InstanceSize,\Witness_0)), \IARGVerifier^{\ROInstance}(1^\Security,0^\InstanceSize) \rangle\\
  \Witness' \gets \WHOutAdversary^\ROInstance(0^\InstanceSize,\IARGVerifierRandomness,\IARGTranscript)\\
  (\IARGProverMessage_{1},\IARGVerifierMessage_{1},\dots,\IARGProverMessage_{\IARGRoundComplexity},\IARGVerifierMessage_{\IARGRoundComplexity}) \coloneq \IARGTranscript
\end{array}
}
\enspace,
\end{equation}
by \cref{lem:how-modifying-oracle-changes-distributions} with $\GenericOracleAlgorithmInputSet = \bits^\IARGVerifierRandomnessComplexity\times\bits^{\IARGProverComm}$, $\GenericPredicate(\ROInstance,\Witness_0,\GenericOracleAlgorithmInput=(\IARGVerifierRandomness,(\IARGProverMessage_{i})_{i\in[\IARGRoundComplexity]})) = \VQueryPredicate{\ROInstance}{(\IARGProverMessage_{i})_{i\in[\IARGRoundComplexity]}}{\IARGVerifierRandomness}{\Witness_0}$, 
\begin{equation*}
\Distribution_{(\ROInstance,\Witness_0)} 
=
\left\{
  \GenericOracleAlgorithmInput=(\IARGVerifierRandomness,(\IARGProverMessage_{i})_{i\in[\IARGRoundComplexity]})
  \;\middle\vert\;
  \begin{array}{l}
  \IARGDecision \xleftarrow{\IARGProverRandomness,\IARGVerifierRandomness,(\IARGProverMessage_{1},\IARGVerifierMessage_{1},\dots,\IARGProverMessage_{\IARGRoundComplexity},\IARGVerifierMessage_{\IARGRoundComplexity})}\langle\IARGProver^{\ROInstance}(1^\Security,(0^n,\Witness_0)),\IARGVerifier^{\ROInstance}(1^\Security,0^n)\rangle
  \end{array}
\right\}
\end{equation*}
for $(\ROInstance,\Witness_0)\in\RandomOracle_{\Security,\InstanceSize}\times(\WitnessDomain\cup\{\bot\})$. Because $\WHOutAdversary$ outputs a valid witness if the verifier makes a query at a location relevant for $\Witness_0$, \Cref{eq:wh-proof-query-exp-for-original-oracle} is upper bounded by \Cref{eq:wh-prob}. By the equality of \Cref{eq:wh-proof-query-exp-for-original-oracle} and \Cref{eq:wh-proof-query-exp-for-modified-oracle}, we conclude that \Cref{eq:wh-proof-query-exp-for-modified-oracle} is at most \Cref{eq:wh-prob}, and, consequently, \Cref{eq:wh-main-lowerbound} holds.
\end{proof}
\doclearpage
\section{Zero-knowledge relativized arguments in the signed oracle model}
\label{sec:zk-constructions}
We construct a zero-knowledge argument with succinct verification for every relation in $\NTIME^{\GeneralOracle}[T]$ (\cref{def:relativized-ntime}) in the signed oracle model.

We authenticate oracle answers using the signature-card approach of \cite{ChiesaT10}. We consider an oracle model in which a trusted party authenticates every oracle answer. Informally, a trusted party holds a signing key and, on a query $\OracleQuery$, returns the oracle answer $\OracleAnswer=\GeneralOracleInstance(\OracleQuery)$ together with a signature on the query--answer pair $(\OracleQuery,\OracleAnswer)$. 

\begin{definition}[Signed oracle]
\label{def:signed-oracle}
Let $\GeneralOracle$ be an oracle and let $\SigScheme=(\SigGen,\SigSign,\SigVerify)$ be a digital signature scheme as in \Cref{def:signature}. For a function $\GeneralOracleInstance\colon\FunDomain\to\FunRange$ with $\KeyQuery\notin\FunDomain$, a verification key $\SigVK$, and signatures $\vect{\Sig}=(\Sig_{\OracleQuery})_{\OracleQuery\in\FunDomain}$, we write $\SignedFunction(\GeneralOracleInstance,\SigVK,\vect{\Sig})$ for the function defined by
\begin{equation*}
\SignedFunction(\GeneralOracleInstance,\SigVK,\vect{\Sig})(\OracleQuery)
\coloneq
\begin{cases}
(\GeneralOracleInstance(\OracleQuery),\Sig_{\OracleQuery}) &\text{if }\OracleQuery\in\FunDomain\\
\SigVK &\text{if }\OracleQuery=\KeyQuery
\end{cases}
\enspace,
\end{equation*}
and call $\GeneralOracleInstance$ the underlying oracle of $\SignedFunction(\GeneralOracleInstance,\SigVK,\vect{\Sig})$. The \defemph{signed oracle} associated with $\GeneralOracle$ and $\SigScheme$, denoted $\SignedOracle$, is the collection $\{\SignedOracle_{\Security,\InstanceSize}\}_{\Security,\InstanceSize\in\N}$, where $\SignedOracle_{\Security,\InstanceSize}$ is the following distribution:
\begin{equation*}
\left\{
\SignedFunction(\GeneralOracleInstance,\SigVK,\vect{\Sig})
\;\middle\vert\;
\begin{array}{l}
\GeneralOracleInstance\gets\GeneralOracle_{\Security,\InstanceSize} \\
(\SigSK,\SigVK)\gets\SigGen(1^\Security)\\
\forall\OracleQuery\in\FunDomain_{\Security,\InstanceSize}\colon\ \Sig_{\OracleQuery}\gets\SigSign(\SigSK,(\OracleQuery,\GeneralOracleInstance(\OracleQuery)))\\
\vect{\Sig}\coloneq(\Sig_{\OracleQuery})_{\OracleQuery\in\FunDomain_{\Security,\InstanceSize}}
\end{array}
\right\}
\enspace.
\end{equation*}
\end{definition}

In practice, the signed oracle is implemented by a trusted party that holds the signing key, has access to the oracle, and lazily signs each query--answer pair it is asked about, reusing the signature for repeated queries. The party performs no other computation, but it \emph{sees the queries}, which in our applications contain private data; it is trusted to sign only correct query--answer pairs and to keep the signing key and the queries private.

For a relativized relation $\Relation_{\GeneralOracle}$, we define its \defemph{signed-oracle lift} $\Relation_{\SignedOracle}$ by preserving the relation of the underlying oracle: for every signed-oracle instance $\SignedOracleInstance$ with underlying oracle $\GeneralOracleInstance$,
\begin{equation*}
\Relation_{\SignedOracleInstance}
\coloneq
\Relation_{\GeneralOracleInstance}
\enspace.
\end{equation*}
Thus, the signatures authenticate oracle answers, but do not change which instance-witness pairs belong to the relation. 

We prove that there exists a zero-knowledge argument for the signed-oracle lift of every relation in $\NTIME^{\GeneralOracle}[T]$.
\begin{theorem}
\label{thm:zk-signed-oracle}
Let $\GeneralOracle$ be an oracle with $\FunRange_{\Security,\InstanceSize}\subseteq\bits^{\leq\poly(\Security,\InstanceSize)}$ and membership in $\FunDomain_{\Security,\InstanceSize}$ decidable in time $\poly(\Security,\InstanceSize)$, let $T\colon\N\to\N$ be a time-constructible function with $T(\InstanceSize)\geq\InstanceSize$ that is computable in time $\poly(\InstanceSize,\log T(\InstanceSize))$, let $\SigScheme$ be a digital signature scheme that is EUF-CMA secure relative to $\GeneralOracle$ (\cref{def:signature}), and let $\SignedOracle$ be the signed oracle associated with $\GeneralOracle$ and $\SigScheme$ (\cref{def:signed-oracle}). If there exists a keyed hash family that is collision resistant relative to $\SignedOracle$ (\cref{def:crh}), then for every relativized relation $\Relation_\GeneralOracle\in\NTIME^{\GeneralOracle}[T]$ there exists a relativized argument $\IARG$ for its signed-oracle lift $\Relation_{\SignedOracle}$ with perfect completeness, soundness error $\negl{\Security}$ against polynomial-time provers for $\InstanceSize,T(\InstanceSize)\leq\poly(\Security)$, $2\Security+1$ rounds, communication complexity $\poly(\Security,\log T(\InstanceSize))$, prover time $\poly(\Security,T(\InstanceSize))$, verifier time $\poly(\Security,\InstanceSize,\log T(\InstanceSize))$, and malicious-verifier zero-knowledge error at most $\Security\cdot2^{-\Security}$ with simulator time $\poly(\Security,T(\InstanceSize),\ZKAdversaryTime)$ and query complexity $\poly(\Security,T(\InstanceSize),\ZKAdversaryQuery)$. The honest prover makes $O(T(\InstanceSize))$ oracle queries, and the honest verifier queries the oracle only for the verification key.
\end{theorem}

The hypotheses are satisfied, for example, by a keyed hash family $\HashFamily$ that is collision resistant relative to the signed oracle of the signature scheme that \cref{cor:digital-signature-exists} builds from $\HashFamily$: collision resistance relative to $\SignedOracle$ implies collision resistance relative to $\GeneralOracle$, since an adversary relative to $\GeneralOracleInstance$ is also an adversary relative to every $\SignedOracleInstance$ with underlying oracle $\GeneralOracleInstance$ (it discards the signatures).

Our construction follows \cite{ChiesaT10}, except that we add zero knowledge. In \cref{sec:kilian-oracle} we construct a zero-knowledge argument of knowledge for the (non-relativized) universal relation that is secure relative to any oracle relative to which a collision-resistant hash family exists; in \cref{sec:signed-oracle} we compile it into a zero-knowledge argument for relativized $\NTIME$ relations in the signed oracle model, which yields \cref{thm:zk-signed-oracle}.

\subsection{A zero-knowledge argument of knowledge for the universal relation}
\label{sec:kilian-oracle}

We apply Kilian's protocol~\cite{Kilian92}, in the vector-commitment formulation of \cite{ChiesaDGSY24}, to the HVZK PCP of \cref{thm:hvzk-PCP-knowledge} and the hiding vector commitment scheme of \cref{thm:vc-from-crh}.

\begin{theorem}
\label{thm:zk-aok-universal}
Let $\GeneralOracle$ be an oracle. If there exists a keyed hash family that is collision resistant relative to $\GeneralOracle$ (\cref{def:crh}), then there exists an interactive argument for the universal relation $\URel$ (\cref{def:universal-relation}) with the following properties, where $(M,\Instance,t)$ denotes the instance.
\begin{itemize}[noitemsep]
  \item Perfect completeness.
  \item Knowledge soundness error $3/4+\negl{\Security}$ relative to $\GeneralOracle$ for $\abs{M},\abs{\Instance},t\leq\poly(\Security)$ and $\poly(\Security)$-time provers, with extraction time $\poly(\Security,t,\KSAdversaryTime)$.
  \item Malicious-verifier zero-knowledge error $2^{-\Security}$ relative to $\GeneralOracle$, with simulator time $\poly(\Security,t,\ZKAdversaryTime)$ and simulator query complexity $\poly(\Security,t,\ZKAdversaryQuery)$.
  \item Communication complexity $\poly(\Security,\log t)$, prover time $\poly(\Security,\abs{M}+\abs{\Instance}+t)$, and verifier time $\poly(\Security,\abs{M}+\abs{\Instance}+\log t)$. The honest prover and verifier make no oracle queries.
\end{itemize}
\end{theorem}
\begin{proof}
We instantiate Kilian's protocol~\cite{Kilian92}, in the vector-commitment formulation of \cite[Construction~4.1]{ChiesaDGSY24}, with the PCP of \cref{thm:hvzk-PCP-knowledge} and the vector commitment scheme of \cref{thm:vc-from-crh}, whose public parameters are sampled by the verifier and sent as its first message (\cref{sec:general-oracle-relativized-args}). Perfect completeness, the communication complexity, and the prover and verifier times follow from the parameters of the PCP and the succinctness of the vector commitment scheme. Knowledge soundness follows from \cite[Theorem~5.1]{ChiesaDGSY24}. Honest-verifier zero knowledge follows from \cite[Section~6.1, Remark~6.10]{IshaiMSX15}. One can apply the guess-and-rewind simulator of \cite{GoldreichMW91} to the honest-verifier simulator to obtain malicious-verifier zero knowledge; instantiating the vector commitment scheme with security parameter $\Security+O(\log\Security+\log t)$ makes the error of this simulator at most $2^{-\Security}$.
\end{proof}

\subsection{Proof of \texorpdfstring{\cref{thm:zk-signed-oracle}}{Theorem~\ref{thm:zk-signed-oracle}}}
\label{sec:signed-oracle}

We show how to generically use a malicious-verifier zero-knowledge argument of knowledge for the universal relation to construct a malicious-verifier zero-knowledge relativized argument for relativized $\NTIME$ relations in the signed oracle model.

\parhead{The signature relation}
Let $\GeneralOracle$ be an oracle, let $\SigScheme=(\SigGen,\SigSign,\SigVerify)$ be a digital signature scheme that is EUF-CMA secure relative to $\GeneralOracle$, and let $\SignedOracle$ be the signed oracle associated with $\GeneralOracle$ and $\SigScheme$, as in \cref{thm:zk-signed-oracle}. Let $\Relation_\GeneralOracle\in\NTIME^{\GeneralOracle}[T]$ and let $M$ be its ND-decider (\cref{def:relativized-ntime}), which runs in time $O(T(\InstanceSize))$. We focus on the instances of the universal relation $\URel$ (\cref{def:universal-relation}) of the form $\Instance_{\mathsf{sig}}\coloneq(M_{\mathsf{sig}},(1^\Security,\Instance,\SigVK),t)$, where $M_{\mathsf{sig}}$ is the machine that, on input $((1^\Security,\Instance,\SigVK),(\Witness,((q_i,a_i,\sigma_i))_{i}))$, runs $M$ on $(\Instance,\Witness)$, answering queries outside $\FunDomain_{\Security,\InstanceSize}$ with the empty string and its $i$-th query in $\FunDomain_{\Security,\InstanceSize}$ with $a_i$ after checking that the query equals $q_i$, checks that $\SigVerify(\SigVK,(q_i,a_i),\sigma_i)=1$ for every $i$, and accepts if and only if $M$ accepts and all checks pass, and $t\coloneq\poly(\Security,T(\InstanceSize))$ is a bound on the running time of $M_{\mathsf{sig}}$. The machine $M_{\mathsf{sig}}$ does not query the oracle: it ensures that the oracle trace is signed, not that its answers agree with $\GeneralOracleInstance$.

\parhead{Construction}
Let $\IARG_{\mathsf{sig}}=(\IARGProver_{\mathsf{sig}},\IARGVerifier_{\mathsf{sig}})$ be a malicious-verifier zero-knowledge argument of knowledge for $\URel$ relative to $\SignedOracle$. We construct $\IARG=(\IARGProver,\IARGVerifier)$ for $\Relation_{\SignedOracle}$ as follows. On common input $(1^\Security,\Instance)$ and prover witness $\Witness$:
\begin{enumerate}[noitemsep]
\item Both $\IARGProver$ and $\IARGVerifier$:
  \begin{enumerate}[nolistsep]
    \item Query $\SignedOracleInstance(\KeyQuery)$ to obtain $\SigVK$.
    \item Set $\Instance_{\mathsf{sig}} \coloneq (M_{\mathsf{sig}},(1^\Security,\Instance,\SigVK),t)$.
  \end{enumerate}
\item $\IARGProver$:
  \begin{enumerate}[nolistsep]
    \item Run $M$ on $(\Instance,\Witness)$, answering its $i$-th query $q_i\in\FunDomain_{\Security,\InstanceSize}$ with $a_i$, where $(a_i,\sigma_i)\coloneq\SignedOracleInstance(q_i)$, and its other queries with the empty string.
    \item Set $\Witness_{\mathsf{sig}}\coloneq(\Witness,((q_i,a_i,\sigma_i))_{i})$.
  \end{enumerate}
\item $\IARGProver$ and $\IARGVerifier$ run $\IARGDecision\gets \langle\IARGProver_{\mathsf{sig}}(1^\Security,\Instance_{\mathsf{sig}},\Witness_{\mathsf{sig}}),\IARGVerifier_{\mathsf{sig}}(1^\Security,\Instance_{\mathsf{sig}})\rangle$.
\item $\IARGVerifier$ outputs $\IARGDecision$.
\end{enumerate}

\parhead{Analysis}
We instantiate $\IARG_{\mathsf{sig}}$ with the argument of \cref{thm:zk-aok-universal} relative to $\SignedOracle$, which exists by the hypothesis of \cref{thm:zk-signed-oracle}, and prove that $\IARG$ has the following properties; \cref{thm:zk-signed-oracle} then follows by sequential repetition.
\begin{itemize}[noitemsep]
  \item Perfect completeness.
  \item Soundness error $3/4+\negl{\Security}$ for $\InstanceSize,T(\InstanceSize)\leq\poly(\Security)$ and $\poly(\Security)$-time provers.
  \item Malicious-verifier zero-knowledge error $2^{-\Security}$, with simulator time $\poly(\Security,T(\InstanceSize),\ZKAdversaryTime)$ and simulator query complexity $\poly(\Security,T(\InstanceSize),\ZKAdversaryQuery)$.
  \item Three rounds, communication complexity $\poly(\Security,\log T(\InstanceSize))$, prover time $\poly(\Security,T(\InstanceSize))$, and verifier time $\poly(\Security,\InstanceSize,\log T(\InstanceSize))$. The honest prover makes $O(T(\InstanceSize))$ oracle queries, one for the verification key and at most one per oracle query of $M$, and the honest verifier queries the oracle once, for the verification key.
\end{itemize}

\parhead{Completeness and efficiency}
Perfect completeness follows from the perfect correctness of $\SigScheme$ and the perfect completeness of $\IARG_{\mathsf{sig}}$. The instance $\Instance_{\mathsf{sig}}$ has length $O(\Security+\InstanceSize+\abs{\SigVK}+\log t)$ and time bound $t=\poly(\Security,T(\InstanceSize))$, so the round complexity, the communication complexity, and the prover and verifier times of $\IARG_{\mathsf{sig}}$ are as claimed; beyond running $\IARGVerifier_{\mathsf{sig}}$, the verifier $\IARGVerifier$ spends $\poly(\Security,\InstanceSize,\log T(\InstanceSize))$ time. Since the honest parties of $\IARG_{\mathsf{sig}}$ make no oracle queries, the honest prover makes $O(T(\InstanceSize))$ oracle queries, one for the verification key and at most one per oracle query of $M$, and the honest verifier queries the oracle once, for the verification key.

\parhead{Soundness}
Fix $\Security,\InstanceSize\in\N$ with $\InstanceSize,T(\InstanceSize)\leq\poly(\Security)$, $\GeneralOracleInstance\in\GeneralOracle_{\Security,\InstanceSize}$, an instance $\Instance\notin\Language(\Relation_{\GeneralOracleInstance})$, and a $\poly(\Security)$-time malicious prover $\Malicious{\IARGProver}$ for $\IARG$. Let $\SignedOracleInstance\gets\SignedOracle_{\Security,\InstanceSize}$ be a signed oracle with underlying oracle $\GeneralOracleInstance$, and consider the experiment
\begin{equation*}
\left\{
\begin{array}{l}
\IARGDecision \xleftarrow{\IARGTranscript} \langle \Malicious{\IARGProver}^{\SignedOracleInstance}(\Instance), \IARGVerifier^{\SignedOracleInstance}(1^\Security,\Instance)\rangle\\
\Witness_{\mathsf{sig}}\gets \KSExtractor^{\SignedOracleInstance,\Malicious{\IARGProver}(\Instance)}(1^\Security,\Instance_{\mathsf{sig}},\IARGTranscript,\IARGDecision)
\end{array}
\right\}
\enspace,
\end{equation*}
where $\KSExtractor$ is the extractor of $\IARG_{\mathsf{sig}}$. Since $\IARGVerifier$ only forwards the messages of $\IARGVerifier_{\mathsf{sig}}$, $\Malicious{\IARGProver}$ is a malicious prover for $\IARG_{\mathsf{sig}}$ on $\Instance_{\mathsf{sig}}$, so by the knowledge soundness of $\IARG_{\mathsf{sig}}$ relative to $\SignedOracle$ (\cref{thm:zk-aok-universal}),
\begin{equation*}
\prob{\IARGDecision=1\,\land\,(\Instance_{\mathsf{sig}},\Witness_{\mathsf{sig}})\notin\URel}\leq3/4+\negl{\Security}
\enspace.
\end{equation*}
We show that the probability that $\IARGDecision=1$ and $(\Instance_{\mathsf{sig}},\Witness_{\mathsf{sig}})\in\URel$ is $\negl{\Security}$, which gives the claimed soundness error. If $\IARGDecision=1$ and $(\Instance_{\mathsf{sig}},\Witness_{\mathsf{sig}})\in\URel$ with $\Witness_{\mathsf{sig}}=(\Witness,((\OracleQuery_i,\OracleAnswer_i,\Sig_i))_{i})$, then $\OracleAnswer_i\neq\GeneralOracleInstance(\OracleQuery_i)$ for some $i$, since otherwise the run of $M$ inside $M_{\mathsf{sig}}$ is the run of $M^{\GeneralOracleInstance}$ on $(\Instance,\Witness)$, which accepts, so $(\Instance,\Witness)\in\Relation_{\GeneralOracleInstance}$; moreover $\SigVerify(\SigVK,(\OracleQuery_i,\OracleAnswer_i),\Sig_i)=1$, while $\SignedOracleInstance$ signs only messages of the form $(\OracleQuery,\GeneralOracleInstance(\OracleQuery))$. Hence $\Sig_i$ is a forgery, and the following EUF-CMA adversary outputs it.
\begin{itemize}[nolistsep]
  \item $\SigAdv^{\GeneralOracleInstance,\SigSign(\SigSK,\cdot)}(1^\Security,\SigVK)$:
  \begin{enumerate}[nolistsep]
    \item Set $\Instance_{\mathsf{sig}} \coloneq (M_{\mathsf{sig}},(1^\Security,\Instance,\SigVK),t)$.
    \item Run the above experiment, answering each oracle call $\SignedOracleInstance(\KeyQuery)$ with $\SigVK$ and each oracle call $\SignedOracleInstance(\OracleQuery)$ with $(\GeneralOracleInstance(\OracleQuery),\SigSign(\SigSK,(\OracleQuery,\GeneralOracleInstance(\OracleQuery))))$, where the signature of each query is requested once and reused for repeated queries.
    \item If $\IARGDecision=0$ or $(\Instance_{\mathsf{sig}}, \Witness_{\mathsf{sig}})\notin \URel$, output $\bot$. Otherwise, output $((\OracleQuery_i,\OracleAnswer_i),\Sig_i)$ for the first $i$ with $\OracleAnswer_i\neq\GeneralOracleInstance(\OracleQuery_i)$.
  \end{enumerate}
\end{itemize}
The answers of $\SigAdv$ to the oracle calls are distributed exactly as $\SignedOracleInstance$, so $\SigAdv$ forges exactly when $\IARGDecision=1$ and $(\Instance_{\mathsf{sig}},\Witness_{\mathsf{sig}})\in\URel$. Since $\SigAdv$ runs in time $\poly(\Security)$ (\cref{thm:zk-aok-universal} bounds the extraction time), this probability is $\negl{\Security}$ by the EUF-CMA security of $\SigScheme$ relative to $\GeneralOracle$.

\parhead{Zero knowledge}
The simulator for $\IARG$ queries $\SignedOracleInstance(\KeyQuery)$, computes $\Instance_{\mathsf{sig}}$ as the honest verifier would, and runs the simulator of $\IARG_{\mathsf{sig}}$ on $\Instance_{\mathsf{sig}}$ with the malicious verifier. Therefore, the zero knowledge of $\IARG$ follows from that of $\IARG_{\mathsf{sig}}$ (\cref{thm:zk-aok-universal}), with the same error and one additional oracle query.

\parhead{Sequential repetition}
Let $\IARG'$ run steps 1 and 2 of $\IARG$ once and then step 3 $\Security$ times sequentially, accepting if and only if every session accepts; the verifier sends the parameters of each session with its last message of the previous session, so $\IARG'$ has $2\Security+1$ rounds. Completeness, communication complexity, verifier time, and the honest parties' oracle queries are as claimed in \cref{thm:zk-signed-oracle}. For soundness, run $\KSExtractor$ after each session on the malicious prover from its state at the beginning of the session. Conditioned on the previous sessions, each session is an execution of $\IARG_{\mathsf{sig}}$, so the probability that every session accepts and no extraction yields a witness for $\Instance_{\mathsf{sig}}$ is at most $(3/4+\negl{\Security})^{\Security}=\negl{\Security}$, while the probability that some extraction yields a witness for $\Instance_{\mathsf{sig}}$ although $\Instance\notin\Language(\Relation_{\GeneralOracleInstance})$ is $\negl{\Security}$, since $\SigAdv$ above, run on all sessions and extractions, then outputs a forgery. Hence $\IARG'$ has soundness error $\negl{\Security}$. By the sequential composition of auxiliary-input zero knowledge~\cite{GoldreichO94}, whose hybrid argument adds the errors of the sessions, $\IARG'$ has malicious-verifier zero-knowledge error at most $\Security\cdot2^{-\Security}$, with simulator time and query complexity $\Security$ times those of $\IARG$. This proves \cref{thm:zk-signed-oracle}.

\begin{remark}[Efficiency]
\label{rem:efficiency}
The construction of this section demonstrates feasibility, so we kept it simple. Replacing the PCP and the vector commitment scheme by a modern hash-based succinct argument makes the construction concretely efficient: for example, Flock~\cite{BunzRW26} proves batched boolean computations with a constant prover overhead over native execution and a verifier whose cost is linear in a single computation but sublinear in the size of the batch, and adding zero knowledge to such systems incurs only a small additional prover cost~\cite{ChiesaFW26}.
\end{remark}


\appendix

\doclearpage
\section{Impossibility of witness hiding for debate}
\label{sec:debate}

We extend \Cref{thm:wh-lower-bound-rom} to debate, the two-prover model of scalable oversight \cite{ICA18,BCIP24}: a debate whose honest opponent is efficient collapses to a single-prover relativized argument with the same guarantees, so the lower bound applies to it.

\parhead{Relativized debates}
Let $\GeneralOracle$ be an oracle. A \textbf{relativized debate} in the oracle model $\GeneralOracle$ for a relativized relation $\Relation_\GeneralOracle$ is a triple of interactive oracle algorithms $\DEBATE=(\DebateProver,\DebateOpponent,\Judge)$: the \emph{prover} $\DebateProver$ receives the witness, the \emph{opponent} $\DebateOpponent$ does not, and the \emph{verifier} $\Judge$ decides. All three have access to the same oracle function $\GeneralOracleInstance\in\GeneralOracle_{\Security,\InstanceSize}$. They interact for $\IARGRoundComplexity$ rounds as in \cite{ICA18,BCIP24}: in round $i$, the prover sends $\IARGProverMessage_i$ and the opponent sends $c_i$, each computed by the party from its input, its randomness, and the messages of the previous rounds; after the last round, the verifier outputs its decision $\IARGDecision\in\bits$, computed from its input, its randomness, and all messages. We write $\IARGDecision\xleftarrow{\IARGProverRandomness_1,\IARGProverRandomness_2,\IARGVerifierRandomness,\IARGTranscript}\langle\DebateProver^\GeneralOracleInstance(1^\Security,(\Instance,\Witness)),\DebateOpponent^\GeneralOracleInstance(1^\Security,\Instance),\Judge^\GeneralOracleInstance(1^\Security,\Instance)\rangle$ for this experiment, where $\IARGProverRandomness_1,\IARGProverRandomness_2,\IARGVerifierRandomness$ are the randomness of the three parties and $\IARGTranscript\coloneq(\IARGProverMessage_i,c_i)_{i\in[\IARGRoundComplexity]}$ is the transcript, and omit the randomness and the transcript when they are irrelevant. The efficiency measures of \Cref{sec:general-oracle-relativized-args} apply to each party: $\IARGVerifierQuery$ and $\IARGVerifierTime$ bound the queries and running time of the verifier, $\IARGProverQuery_1$ and $\IARGProverTime_1$ those of the honest prover, and $\IARGProverQuery_2$ and $\IARGProverTime_2$ those of the honest opponent.

\begin{remark}[Debate with a witness]
\label{rem:debate-witness}
\cite{BCIP24} also consider debates for relations with a witness. In their model, the prover announces the witness at the start of the debate, so the opponent and the verifier receive it. This is the natural choice when the witness is a solution that the verifier should learn, but it rules out any privacy, since the verifier's view contains the witness. We therefore give the witness to the prover only.
\end{remark}

\begin{remark}[Bounded adversaries]
\label{rem:debate-bounded}
\cite{BCIP24} require completeness against every opponent and soundness against every cheating prover, both computationally unbounded. We require them only against query- and time-bounded opponents and cheating provers, as for arguments (\cref{def:rel-arg-adaptive-soundness-gom}), and the errors in \cref{def:debate-completeness,def:debate-soundness} are functions of these bounds. Since the lower bound rules out debates under these weaker requirements, it rules out debates under the requirements of \cite{BCIP24} as well.
\end{remark}

Completeness must hold against every opponent and soundness against every cheating prover, as in \cite{BCIP24}.

\begin{definition}[Completeness]
\label{def:debate-completeness}
$\DEBATE$ has \defemph{completeness error $\IARGCompleteness$} if for every security parameter $\Security\in\N$, instance size bound $\InstanceSize\in\N$, instance $\Instance\in\bits^\InstanceSize$, adversary $\IARGCompletenessAdv$, and $\GenericQueryBound$-query $\GenericTimeBound$-time opponent $\Malicious{\IARGProver}_2$,
\begin{equation*}
\prob{
\begin{array}{l}
  \IARGDecision=0 \\
  \land\, (\Instance,\Witness)\in\Relation_\GeneralOracleInstance
\end{array}
\;\middle\vert\;
\begin{array}{l}
  \GeneralOracleInstance\gets {\GeneralOracle_{\Security,\InstanceSize}}\\
  \Witness\gets \IARGCompletenessAdv^\GeneralOracleInstance(\Instance)\\
  \IARGDecision\gets \langle \DebateProver^\GeneralOracleInstance(1^\Security,(\Instance,\Witness)), \Malicious{\IARGProver}_2^\GeneralOracleInstance(\Instance), \Judge^\GeneralOracleInstance(1^\Security,\Instance)\rangle
\end{array}
}
\leq \IARGCompleteness(\Security,\InstanceSize,\GenericQueryBound,\GenericTimeBound)
\enspace.
\end{equation*}
\end{definition}

\begin{definition}[Soundness]
\label{def:debate-soundness}
$\DEBATE$ has \defemph{soundness error $\IARGSoundness$} if for every security parameter $\Security\in\N$, instance size bound $\InstanceSize\in\N$, instance $\Instance\in\bits^\InstanceSize$, and $\GenericQueryBound$-query $\GenericTimeBound$-time cheating prover $\Malicious{\IARGProver}_1$,
\begin{equation*}
\prob{
\begin{array}{l}
  \IARGDecision=1\\
  \land\,\Instance\notin \Language(\Relation_\GeneralOracleInstance)
\end{array}
\;\middle\vert\;
\begin{array}{l}
  \GeneralOracleInstance\gets {\GeneralOracle_{\Security,\InstanceSize}}\\
  \IARGDecision\gets \langle \Malicious{\IARGProver}_1^\GeneralOracleInstance(\Instance), \DebateOpponent^\GeneralOracleInstance(1^\Security,\Instance), \Judge^\GeneralOracleInstance(1^\Security,\Instance)\rangle
\end{array}
}\leq \IARGSoundness(\Security,\InstanceSize,\GenericQueryBound,\GenericTimeBound)\enspace.
\end{equation*}
\end{definition}

Privacy in a debate must protect the witness from the opponent and the verifier together, since the two may cooperate. We require the weakest such guarantee, in the spirit of \Cref{def:wh}: the joint view of the honest opponent and the honest verifier must not allow recovering a witness. The impossibility below therefore applies to every stronger requirement, such as hiding against a malicious opponent and verifier.

\begin{definition}[Witness hiding]
\label{def:debate-wh}
$\DEBATE$ has \defemph{witness-hiding error $\IARGWitnessHiding$} if for every security parameter $\Security\in\N$, instance size bound $\InstanceSize\in\N$, instance $\Instance\in\bits^\InstanceSize$, adversary $\WHSampleAdversary$, and $\WHAdversaryQuery$-query $\WHAdversaryTime$-time adversary $\WHOutAdversary$,
\begin{equation*}
\prob{
\begin{array}{l}
(\Instance,\Witness')\in\Relation_\GeneralOracleInstance\\
\land\, (\Instance,\Witness)\in\Relation_\GeneralOracleInstance
\end{array}
\;\middle\vert\;
\begin{array}{l}
  \GeneralOracleInstance\gets {\GeneralOracle_{\Security,\InstanceSize}} \\
  \Witness \gets \WHSampleAdversary^\GeneralOracleInstance(\Instance) \\
  \IARGDecision \xleftarrow{\IARGProverRandomness_1,\IARGProverRandomness_2,\IARGVerifierRandomness,\IARGTranscript} \langle \DebateProver^{\GeneralOracleInstance}(1^\Security,(\Instance,\Witness)), \DebateOpponent^{\GeneralOracleInstance}(1^\Security,\Instance), \Judge^{\GeneralOracleInstance}(1^\Security,\Instance) \rangle\\
  \Witness' \gets \WHOutAdversary^{\GeneralOracleInstance}(\Instance,\IARGProverRandomness_2,\IARGVerifierRandomness,\IARGTranscript)
\end{array}
}
\leq
\IARGWitnessHiding(\Security, \InstanceSize, \WHAdversaryQuery,\WHAdversaryTime)
\enspace.
\end{equation*}
\end{definition}

\begin{definition}[Relativized $\DEBATE$ in an oracle model]
\label{def:rel-debate-gom}
Let $\GeneralOracle$ be an oracle. The complexity class
\begin{equation*}
\DEBATE^{\GeneralOracle}
\left[
\begin{array}{l l l l}
  \emph{\textrm{completeness error}}	& 	\IARGCompleteness & = &\IARGCompleteness(\Security,n,\mathsf{q}_{\Malicious{\IARGProver}_2},\mathsf{t}_{\Malicious{\IARGProver}_2})\\
  \emph{\textrm{soundness error}}	&	\IARGSoundness & = & \IARGSoundness(\Security,n,\mathsf{q}_{\Malicious{\IARGProver}_1},\mathsf{t}_{\Malicious{\IARGProver}_1})\\
  \emph{\textrm{witness-hiding error}}	&	\IARGWitnessHiding & = & \IARGWitnessHiding(\Security,n,\WHAdversaryQuery,\WHAdversaryTime)\\
  \emph{\textrm{verifier query bound}} & \IARGVerifierQuery & = & \IARGVerifierQuery(\Security,n)\\
  \emph{\textrm{verifier running time}}	&	\IARGVerifierTime & = & \IARGVerifierTime(\Security,n)\\
  \emph{\textrm{honest prover query bound}}	&	\IARGProverQuery_1 & = & \IARGProverQuery_1(\Security,n)\\
  \emph{\textrm{honest prover time bound}}	&	\IARGProverTime_1 & = & \IARGProverTime_1(\Security,n)\\
  \emph{\textrm{honest opponent query bound}}	&	\IARGProverQuery_2 & = & \IARGProverQuery_2(\Security,n)\\
  \emph{\textrm{honest opponent time bound}}	&	\IARGProverTime_2 & = & \IARGProverTime_2(\Security,n)\\
\end{array}
\right]
\end{equation*}
is the set of relativized relations $\Relation_{\GeneralOracle}$ that admit a relativized debate $(\DebateProver,\DebateOpponent,\Judge)$ with the listed errors and efficiency measures. A parameter that is omitted is unconstrained.
\end{definition}

\parhead{Folding the opponent into the verifier}
Given a relativized debate, we obtain a single-prover relativized argument by letting the verifier run the honest opponent internally.

\begin{lemma}
\label{lem:debate-folding}
Let $\DEBATE=(\DebateProver,\DebateOpponent,\Judge)$ be a relativized debate for $\Relation_\GeneralOracle$ with errors and efficiency measures as in \cref{def:rel-debate-gom}. There exists a relativized argument $\IARG=(\IARGProver,\IARGVerifier)$ for $\Relation_\GeneralOracle$ with the following properties.
\begin{itemize}[noitemsep]
  \item Completeness error $\IARGCompleteness(\Security,\InstanceSize,\IARGProverQuery_2,\IARGProverTime_2)$, soundness error $\IARGSoundness$, and witness-hiding error $\IARGWitnessHiding$.
  \item Verifier query bound $\IARGVerifierQuery+\IARGProverQuery_2$ and verifier running time $O(\IARGVerifierTime+\IARGProverTime_2)$.
  \item Honest prover query bound $\IARGProverQuery_1$ and honest prover time bound $\IARGProverTime_1$.
\end{itemize}
\end{lemma}

\begin{proof}
Let $\IARGProver\coloneq\DebateProver$, and let $\IARGVerifier$, with randomness $(\IARGProverRandomness_2,\IARGVerifierRandomness)$, answer the $i$-th prover message $\IARGProverMessage_i$ with $c_i\coloneq\DebateOpponent^\GeneralOracleInstance(1^\Security,\Instance,(\IARGProverMessage_j,c_j)_{j<i};\IARGProverRandomness_2)$, which does not depend on $\IARGProverMessage_i$, and after the last round output the decision \\
$\Judge^\GeneralOracleInstance(1^\Security,\Instance,(\IARGProverMessage_j,c_j)_{j\in[\IARGRoundComplexity]};\IARGVerifierRandomness)$. The queries and running time of $\IARGVerifier$ are those of $\DebateOpponent$ and $\Judge$ combined, and those of $\IARGProver$ are those of $\DebateProver$. An interaction of $\IARG$ is an execution of $\DEBATE$ with the honest opponent, with the same randomness and transcript, and a cheating prover for $\IARG$ sees, before sending $\IARGProverMessage_i$, exactly the messages of the previous rounds. Hence:
\begin{itemize}[noitemsep]
  \item the completeness experiment of $\IARG$ (\Cref{def:rel-arg-adaptive-completeness-gom}) is the completeness experiment of $\DEBATE$ (\Cref{def:debate-completeness}) with the opponent $\Malicious{\IARGProver}_2\coloneq\DebateOpponent$, which makes at most $\IARGProverQuery_2$ queries and runs in time $\IARGProverTime_2$;
  \item the soundness experiment of $\IARG$ (\Cref{def:rel-arg-adaptive-soundness-gom}) against a cheating prover is the soundness experiment of $\DEBATE$ (\Cref{def:debate-soundness}) against the same cheating prover;
  \item in the witness-hiding experiment of $\IARG$ (\Cref{def:wh}), the adversary receives the instance, the verifier randomness $(\IARGProverRandomness_2,\IARGVerifierRandomness)$, and the transcript, which is exactly the input $(\Instance,\IARGProverRandomness_2,\IARGVerifierRandomness,\IARGTranscript)$ of the adversary in \Cref{def:debate-wh}, sampled from the same distribution.
\end{itemize}
The three error bounds follow.
\end{proof}

\parhead{The lower bound for debate}
Combining \Cref{lem:debate-folding} with \Cref{thm:wh-lower-bound-rom} gives the following.

\begin{corollary}
\label{cor:wh-debate}
Let $t\colon \N\to \N$ be a time-constructible function with $t(n)\geq n$ and $\RandomOracle$ be the random oracle over $((\bits^{t(n)+\lceil\log t(n)\rceil+1},\bits^\Security))_{\Security,\InstanceSize\in\N}$.
{\small
\begin{equation*}
\NTIME^{\RandomOracle}  \left[t\right]\not\subseteq\;
\DEBATE^{\RandomOracle}
\left[
\begin{array}{l l l l}
  \emph{\textrm{completeness error}}	& 	\IARGCompleteness & = & \begin{cases}
    1/10 &
      \mathsf{q}_{\Malicious{\IARGProver}_2}=\poly(\Security,t(n))\enspace, \\&\mathsf{t}_{\Malicious{\IARGProver}_2}=\poly(\Security,t(n))\\
    1 & \text{otherwise}\\
  \end{cases}  \\
  \emph{\textrm{soundness error}}	&	\IARGSoundness & = & \begin{cases}
    1/10 &
      \mathsf{q}_{\Malicious{\IARGProver}_1}=\poly(\Security,t(n))\enspace, \\&\mathsf{t}_{\Malicious{\IARGProver}_1}=\poly(\Security,t(n))\\
    1 & \text{otherwise}\\
  \end{cases}  \\
  \emph{\textrm{witness-hiding error}}	&	\IARGWitnessHiding & = & \begin{cases}
    1/10 & \WHAdversaryQuery=\poly(\Security,t(n))\enspace,
    \\&\WHAdversaryTime=\poly(\Security,t(n))\\
    1 & \text{otherwise}\\
  \end{cases}  \\
  \emph{\textrm{verifier query bound}}	&	\IARGVerifierQuery & = & \poly(\Security,t(n))\\
  \emph{\textrm{verifier running time}}	&	\IARGVerifierTime & = & \poly(\Security,t(n))\\
  \emph{\textrm{honest prover query bound}}	&	\IARGProverQuery_1 & = & \poly(\Security,t(n))\\
  \emph{\textrm{honest prover time bound}}	&	\IARGProverTime_1 & = & \poly(\Security,t(n))\\
  \emph{\textrm{honest opponent query bound}}	&	\IARGProverQuery_2 & = & \poly(\Security,t(n))\\
  \emph{\textrm{honest opponent time bound}}	&	\IARGProverTime_2 & = & \poly(\Security,t(n))\\
\end{array}
\right]\enspace.
\end{equation*}
}
\end{corollary}

\begin{proof}
If a relation in $\NTIME^{\RandomOracle}[t]$ had a relativized debate with the parameters in the corollary, then \Cref{lem:debate-folding} would give a relativized argument for it with the parameters in \Cref{thm:wh-lower-bound-rom}, contradicting that theorem.
\end{proof}

\begin{remark}
\label{rem:debate-scope}
The proof of \Cref{lem:debate-folding} uses completeness only against the honest opponent, so \Cref{cor:wh-debate} holds even for debates whose completeness is required only against the honest opponent. Since the verifier of \cref{lem:debate-folding} runs the honest opponent, the lemma requires the honest opponent to be efficient, as in doubly-efficient debate \cite{BCIP24}; a debate whose honest opponent runs in more than polynomial time is not covered. Finally, \Cref{cor:wh-debate} concerns witness hiding; zero knowledge against the opponent and the verifier implies witness hiding up to the probability of finding a witness from the instance alone, which is negligible for the relation of \Cref{def:relation-rom-ntime}. \Cref{lem:zk-implies-wh} proves this implication for arguments, and the same proof applies to debates.
\end{remark}

\doclearpage
\clearpage
\iffull

\doclearpage
\section*{Acknowledgments}
\label{sec:acknowledgements}
This work supported in part by the Ethereum Foundation and by the Advanced Research and Invention Agency (ARIA).
\fi

\section*{AI use statement}
\label{sec:ai-use}

We used generative AI tools to help write some parts of the paper. The introduction and the technical overview (\Cref{sec:introduction,sec:techniques}) were written by the authors, with AI assistance to improve the language and to find grammatical mistakes. The lower bound and its proof (\Cref{sec:wh-lower-bound}) were written entirely by the authors, apart from typo and spelling checks. For the extension to debate (\Cref{sec:debate}) and the construction with signed oracles (\Cref{sec:zk-constructions}), the ideas originated from the authors; the first drafts of the proofs were produced by human-guided AI generation, and the current versions were modified and checked by the authors to make sure that they match the intended arguments. Additionally, we used generative AI tools to search for and summarize relevant literature, to prepare the figures, and to obtain feedback on drafts. We have reviewed all AI-assisted work: every reference identified by an LLM was checked manually by at least one of the authors, and all AI-assisted text was revised by the authors. We take responsibility for the final content of this work, including text, claims, and artifacts produced with the aid of generative AI.

\iffull
\printbibliography
\fi
\end{document}